\documentclass[sigconf,nonacm]{acmart}
\renewcommand\footnotetextcopyrightpermission[1]{}
\acmConference{}{}{}
\acmBooktitle{}
\usepackage{xcolor}
\usepackage{float}
\usepackage{wrapfig}
\usepackage{bbm}
\usepackage{algorithm}
\usepackage{algorithmic}
\usepackage[textsize=tiny,textwidth=.98\marginparwidth]{todonotes}
\usepackage{subcaption}

\usepackage{enumitem}
\usepackage{amsmath}
\DeclareMathOperator*{\argmin}{argmin}
\usepackage{dsfont}
\usepackage{xspace}
\newcommand{\OurMethod}{REACT\xspace}
\newcommand{\OurMethods}{{\OurMethod}'s\xspace}
\usepackage[normalem]{ulem}

\usepackage{xcolor}
\definecolor{myDarkGreen}{rgb}{0.0, 0.5, 0.0}
\definecolor{deepGreen}{HTML}{006400}
\usepackage{dblfloatfix}  
\usepackage{placeins}     
\usepackage{cuted}
\usepackage{caption}
\usepackage{float}

\begin{document}

\title{Relaxed Efficient Acquisition of Context and Temporal Features}


\author{%
  Yunni Qu$^{1}$,~
  Dzung Dinh$^{1}$,~
  Grant King$^{2}$,~
  Whitney Ringwald$^{3}$,~ 
  Bing Cai Kok$^{1}$,~\\
  Kathleen Gates$^{1}$,~
  Aidan Wright$^{2}$,~
  Junier Oliva$^{1}$\\[0.5em]
}
\affiliation{%
  \institution{%
    $^{1}$University of North Carolina at Chapel Hill, Chapel Hill, NC, USA \\
    $^{2}$University of Michigan, Ann Arbor, MI, USA \\
    $^{3}$University of Minnesota Twin Cities, Minneapolis, MN, USA \\[0.5em]
    \normalfont\small
    quyunni@cs.unc.edu \quad ddinh@cs.unc.edu \quad grking@umich.edu \quad wringwal@umn.edu \quad
    bingcai@unc.edu \\
     kmgates@unc.edu \quad aidangcw@umich.edu \quad joliva@cs.unc.edu
  }
  \country{}
}

\begin{abstract}
In many biomedical applications, measurements are not freely available at inference time: each laboratory test, imaging modality, or assessment incurs financial cost, time burden, or patient risk. Longitudinal active feature acquisition (LAFA) seeks to optimize predictive performance under such constraints by adaptively selecting measurements over time, yet the problem remains inherently challenging due to temporally coupled decisions (missed early measurements cannot be revisited, and acquisition choices influence all downstream predictions). Moreover, real-world clinical workflows typically begin with an initial onboarding phase, during which relatively stable contextual descriptors (e.g., demographics or baseline characteristics) are collected once and subsequently condition longitudinal decision-making. Despite its practical importance, the efficient selection of onboarding context has not been studied jointly with temporally adaptive acquisition. We therefore propose REACT (Relaxed Efficient Acquisition of Context and Temporal features), an end-to-end differentiable framework that simultaneously optimizes (i) selection of onboarding contextual descriptors and (ii) adaptive feature--time acquisition plans for longitudinal measurements under cost constraints. REACT employs a Gumbel--Sigmoid relaxation with straight-through estimation to enable gradient-based optimization over discrete acquisition masks, allowing direct backpropagation from prediction loss and acquisition cost. Across real-world longitudinal health and behavioral datasets, REACT achieves improved predictive performance at lower acquisition costs compared to existing longitudinal acquisition baselines, demonstrating the benefit of modeling onboarding and temporally coupled acquisition within a unified optimization framework.
\end{abstract}



\renewcommand{\shortauthors}{}
\maketitle

\section{Introduction}


Longitudinal data—from digital health monitoring to behavioral assessments—create new opportunities for early detection and personalized decision-making \cite{shiffman2008ecological, swinckels2024use}. In deployment, however, the challenge is not only prediction but also \emph{resource-constrained measurement at inference time}: acquiring every variable at every time point is often costly, burdensome, and unnecessary. This motivates frameworks that \textbf{adaptively decide what to measure under budget constraints}, explicitly trading off \emph{predictive performance} and \emph{acquisition cost} rather than assuming fully observed inputs~\cite{saar2009active, sheng2006feature, yin2020reinforcement}. In practice, such methods better reflect real longitudinal workflows by prioritizing the most informative measurements at each occasion, deferring costly assessments until warranted, and stopping further collection when its expected value is low.

Consider a mobile behavioral health application that provides just-in-time adaptive support for risk reduction. \emph{During onboarding}, users report contextual descriptors such as demographics, clinical history, and baseline psychometric assessments. \emph{After enrollment}, the application administers ecological momentary assessments (EMAs) at scheduled intervals to capture time-varying signals such as affect, cravings, stressors, and social context for short-term risk forecasting \cite{shiffman2008ecological}. In both phases, however, exhaustive data collection is costly and burdensome: long onboarding surveys can \textbf{deter engagement and limit adoption} ~\cite{aiyegbusikey, rolstad2011response}, while frequent or repetitive EMAs can \textbf{induce fatigue}, \textbf{reduce adherence}, and \textbf{degrade response quality} ~\cite{cook2025understanding}. Rather than collecting all onboarding descriptors and EMA measurements, our approach learns an adaptive acquisition policy that decides \textbf{what to measure and when}, explicitly balancing predictive accuracy against user burden and measurement cost over time.

\begin{figure*}[t]
\centering
\includegraphics[width=0.85\linewidth]{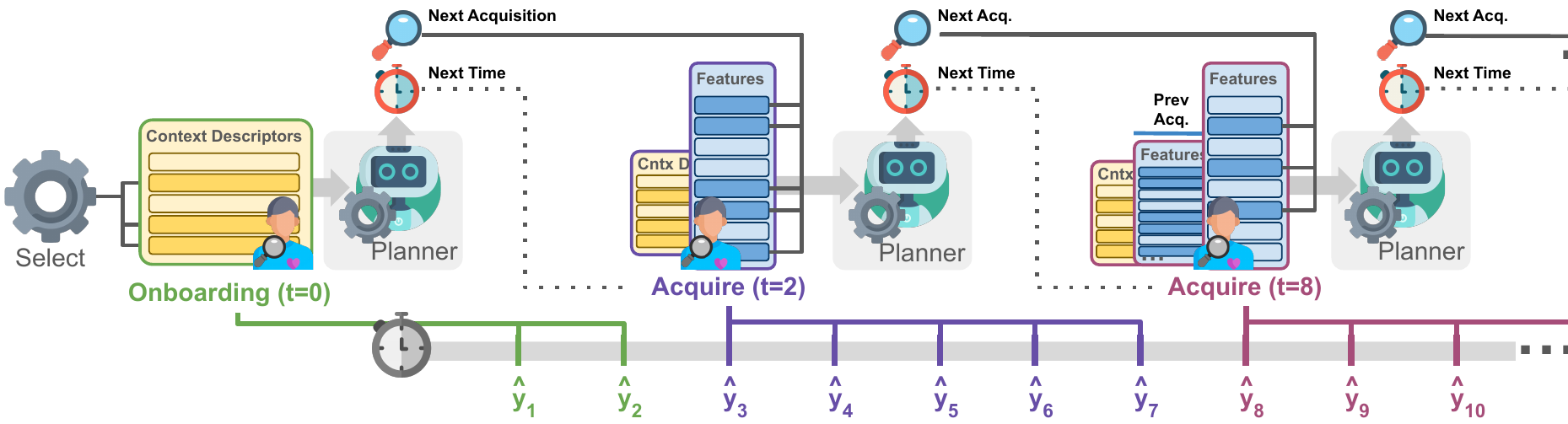}
\caption{\textbf{Overview of REACT.} REACT first performs a one-time onboarding acquisition of contextual descriptors at $t=0$. Shaded feature entries denote acquired measurements. Using acquired context and prior longitudinal observations, the planner then adaptively decides both \emph{when} to acquire next and \emph{which} features to obtain. Between acquisition steps, the model predicts temporal labels without collecting additional measurements, yielding personalized, nonuniform acquisition trajectories.}
\label{fig:highlevel}
\end{figure*}

This same two-stage structure appears more broadly across healthcare onboarding and follow-up. Intake gathers contextual descriptors such as demographics, family history, comorbidities, medication lists, and baseline screening instruments, while subsequent care relies on \emph{longitudinal data collection} tailored to the evolving clinical picture, including follow-up labs, imaging, symptom inventories, and specialist referrals. Exhaustive onboarding can increase administrative burden, prolong visits, raise costs, and reduce completion or response quality ~\cite{aiyegbusikey, rolstad2011response}. Likewise, repeatedly acquiring all possible follow-up measurements is costly, time-intensive, and strains limited clinical resources, including staff time, imaging capacity, and diagnostic equipment. More generally, many healthcare and bioinformatic applications involve one-time acquisition of relatively stable contextual descriptors followed by longitudinal measurement decisions over time. These settings motivate methods that prioritize the most informative and cost-effective acquisitions across both phases.\looseness-1

To address this problem, we propose \textbf{\OurMethod} (\textbf{R}elaxed \textbf{E}fficient \textbf{A}cquisition of \textbf{C}ontext and \textbf{T}emporal features), a relaxation-based framework for longitudinal active feature acquisition (LAFA) that makes cost-aware acquisition decisions for both \emph{a priori contextual descriptors} and \emph{temporally acquired features}. Our contributions are:
\begin{itemize}[leftmargin=*,topsep=0pt, noitemsep, wide]
\item \textbf{Formalizing Onboarding + Longitudinal AFA.}
We introduce a practical LAFA formulation that explicitly distinguishes one-time contextual descriptors acquired at onboarding from sequential, time-indexed measurements acquired during follow-up, better reflecting real-world clinical workflows.
\item \textbf{Joint Context--Temporal Acquisition Learning.}
We develop a unified policy that jointly learns (i) which contextual descriptors to acquire upfront and (ii) a structured \emph{feature\,$\times$\,time} acquisition plan over the longitudinal horizon, optimizing predictive performance under explicit cost constraints.
\item \textbf{Differentiable Discrete Acquisition via Relaxation.}
We enable end-to-end training of discrete acquisition decisions using Gumbel--Sigmoid relaxations ~\cite{Maddison2016TheCD} with straight-through gradients, allowing gradient-based optimization of both the acquisition policy and predictive model with a self-iterative training procedure.
\item \textbf{Empirical Accuracy--Cost Gains.}
Across real-world longitudinal datasets, we demonstrate consistent improvements in the accuracy--cost tradeoff relative to competitive baselines.
\end{itemize}

\section{Related Work}
\label{sec:related_works}

Existing machine learning approaches address important aspects of cost-aware feature acquisition, but key gaps remain for longitudinal settings. Classical feature selection typically operates in a static regime, selecting a population-level subset of variables rather than learning subject-specific acquisition strategies that adapt to previously observed temporal values \cite{miao2016survey, venkatesh2019review, khaire2022stability}. Likewise, active learning focuses primarily on acquiring labels during training, rather than acquiring \emph{features} at \emph{inference} time \cite{houlsby2011bayesian, konyushkova2017learning, mackay1992information}. To position \OurMethod, we review prior work on test-time feature acquisition and highlight limitations in handling the two-phase structure of contextual onboarding followed by longitudinal monitoring.

\subsection{Active Feature Acquisition}
Active Feature Acquisition (AFA) \cite{saar2009active, sheng2006feature} studies test-time prediction when features are not freely available and each measurement incurs a cost. The goal is to sequentially select which features to acquire for a given instance, adapting decisions to previously observed values. Many recent approaches cast AFA as a reinforcement learning problem \cite{shim2018joint, yin2020reinforcement, janisch2020classification}, but RL-based training can be difficult to optimize due to the large action space and challenging credit assignment \cite{li2021active, valancius2024acquisition}. Some methods assuage issues with RL by using generative surrogates to impute missing features and score candidate acquisitions \cite{li2021active}, or by relying on greedy acquisition rules \cite{covert2023learning, ma2018eddi, gong2019icebreaker}, which may fail to capture interactions among feature groups whose utility depends on joint acquisition. To address these limitations, \citet{valancius2024acquisition} propose a non-parametric oracle-based method, \citet{norcliffe2025stochastic} optimize feature acquisition in a stochastic latent space via an expected gradient-based objective, and \citet{ghosh2023difa} use a differentiable policy for end-to-end training of both the acquisition policy and predictor. However, conventional AFA assumes that features are static once acquired and does not model measurements that evolve over time.

\subsection{Longitudinal Active Feature Acquisition}
Longitudinal active feature acquisition (LAFA) extends AFA by requiring the agent to decide not only \emph{what} to acquire, but also \emph{when} to acquire it \cite{kossen2022active, saar2009active}. This introduces additional challenges, since missed measurements at earlier time points may become permanently unavailable. Several recent works cast LAFA as a Markov Decision Process (MDP) \cite{qin2024risk, kossen2022active}. For example, ASAC \cite{yoon2019asac} uses an actor--critic architecture to jointly learn acquisition and prediction, while \citet{qin2024risk} study continuous-time acquisition policies for timely prediction of adverse outcomes. Related approaches also optimize acquisition timing, but may assume that all measurements available at a selected time point are acquired together \cite{nguyen2024active}. As in these sequential formulations, we consider a finite decision horizon with a bounded number of acquisition opportunities.

Although the MDP formulation is natural, RL-based LAFA can be difficult to optimize in practice. The action space is combinatorial, since the policy must jointly determine \emph{which} features to acquire and \emph{when}. The state evolves as new measurements are collected, further complicating policy learning. In addition, supervision is typically provided only through downstream predictive performance, creating a difficult credit-assignment problem for individual acquisitions and decision times \cite{li2021active}.

Beyond these optimization challenges, prior LAFA formulations generally treat all features uniformly and do not explicitly model the common two-phase workflow in which relatively stable contextual descriptors are acquired once at onboarding and then inform subsequent temporal acquisition decisions. In contrast, we study a practical LAFA setting that separates \emph{a priori} contextual descriptors from temporally acquired measurements. We then develop a relaxation-based framework, \OurMethod, that avoids standard RL training by using Gumbel-Sigmoid relaxations with straight-through gradients to enable end-to-end optimization of discrete acquisition decisions.\looseness-1

\section{Method}
In longitudinal active feature acquisition (LAFA), an agent must sequentially choose which measurements to obtain over time, \emph{trading predictive accuracy against acquisition cost}. We propose \OurMethod, an end-to-end framework that jointly learns (i) a one-time selection of onboarding context and (ii) a policy for patient-personalized longitudinal acquisitions. 
Rather than relying on standard reinforcement learning, \OurMethod uses Gumbel-Sigmoid relaxation with straight-through gradients \cite{Maddison2016TheCD} to enable differentiable optimization of discrete acquisition decisions. The model comprises three jointly trained components—a global \emph{Context Selector}, an adaptive \emph{Longitudinal Planner}, and a \emph{Predictor}—optimized under a unified objective that balances prediction loss and acquisition cost. We next formalize the LAFA setting and describe the optimization of each component.\looseness-1

\subsection{Problem Formulation}
\label{sec:formulation}
We consider a training dataset $\mathcal{D}=\{(s^{(i)}, x^{(i)}, y^{(i)})\}_{i=1}^{N}$, where $s^{(i)} \in \mathbb{R}^{d_s}$ denotes the onboarding contextual descriptors for instance $i$, $x^{(i)} \in \mathbb{R}^{T \times d}$ denotes its temporal features across $T$ discrete timesteps, and $y^{(i)}=(y^{(i)}_1,\ldots,y^{(i)}_T)$ denotes the corresponding target sequence with $y^{(i)}_t \in \{1,\ldots,C\}$. For brevity, we omit the superscript $(i)$ when discussing a single instance. While alternative masking schemes could be used---e.g., by passing acquisition masks explicitly to downstream models---we adopt simple elementwise masking for notational convenience. We now describe the iterative acquisition process, which begins with onboarding at $t=0$ and continues until the planner terminates, $\varnothing$, or the final timepoint is reached.\looseness-1

\paragraph{\textbf{Onboarding and First Temporal Acquisition}}
At onboarding (see Fig.~\ref{fig:highlevel}), \OurMethod's Context Selector outputs a static, non-personalized binary mask $m_s \in \{0,1\}^{d_s}$ specifying which contextual descriptors to acquire at cost, yielding $\tilde{s}=m_s \odot s$, where $\odot$ denotes element-wise multiplication. Conditioned on the observed values in $\tilde{s}$, the planner then selects both the next acquisition timepoint, $t_{\mathrm{next}}$, and the temporal feature mask $m_{\mathrm{next}} \in \{0,1\}^{d}$ for the first temporal acquisition. Thus, the first temporal acquisition is personalized through the instance's observed contextual descriptors. For any $t < t_{\mathrm{next}}$, predictions $\hat{y}_t$ are made using only $\tilde{s}$. When $t=t_{\mathrm{next}}$, the selected temporal features are acquired at cost, yielding $\tilde{x}_{t_{\mathrm{next}}}=m_{\mathrm{next}} \odot x_{t_{\mathrm{next}}}$, and prediction proceeds using both $\tilde{s}$ and $\tilde{x}_{t_{\mathrm{next}}}$. The planner then repeats this process, iteratively selecting subsequent acquisition times and feature subsets as described below.\looseness-1

\paragraph{\textbf{Iterative Acquisition Process}}
At each decision step, the planner selects either the next acquisition $(t_{\mathrm{next}}, m_{\mathrm{next}})$ or termination $\varnothing$ in a dynamic, instance-specific manner based on the information acquired so far (see Fig.~\ref{fig:highlevel}). Specifically, it conditions on the contextual descriptors $\tilde{s}$ and the masked temporal history
\[
\mathcal{H}_t = (\tilde{x}_1,\dots,\tilde{x}_t,\mathbf{0},\dots,\mathbf{0}) \in \mathbb{R}^{T\times d},
\]
where unacquired or skipped timepoints remain zero. Thus, \OurMethods planner $\pi$ maps
\[
\pi(\mathcal{H}_t,\tilde{s},t) \rightarrow (t_{\mathrm{next}}, m_{\mathrm{next}})
\quad \text{or} \quad
\pi(\mathcal{H}_t,\tilde{s},t) \rightarrow \varnothing.
\]
As before, for all $t' < t_{\mathrm{next}}$ (or all remaining $t' \leq T$ if the planner terminates), predictions $\hat{y}_{t'}$ are made using only the information acquired so far, namely $(\mathcal{H}_t,\tilde{s})$. If the planner does not terminate, time advances to $t \leftarrow t_{\mathrm{next}}$, the selected temporal features are acquired, yielding $\tilde{x}_{t_{\mathrm{next}}} = m_{\mathrm{next}} \odot x_{t_{\mathrm{next}}}$, and $\hat{y}_{t_{\mathrm{next}}}$ is predicted from the updated history
\[
\mathcal{H}_{t_{\mathrm{next}}} = (\tilde{x}_1,\dots,\tilde{x}_{t_{\mathrm{next}}},\mathbf{0},\dots,\mathbf{0})
\]
together with $\tilde{s}$. The process then repeats.

\paragraph{\textbf{Prediction Model}}
At timestep $t$, a predictor network $f_\phi :  \mathbb{R}^{T \times d} \times \mathbb{R}^{d_s} \times \{1,\dots,T\} \rightarrow \mathbb{R}^C$ takes the temporal history $\mathcal{H}_t$, acquired context $\tilde{s}$, and target timesteps $t'\geq t$ as input, and outputs class probabilities
\begin{equation}
\hat{y}_{t'} = f_\phi(\mathcal{H}_t, \tilde{s}, t').
\label{eq:pred}
\end{equation}

\subsection{The \OurMethod Objective}
Our goal is to acquire contextual and temporal features cost-efficiently while maintaining predictive accuracy and improving the information available for future acquisition decisions. Thus, an acquisition is valuable not only for its immediate predictive benefit, but also for how it improves later planning and prediction.

To this end, we directly train a context selector and planner to output binary acquisition masks that optimize the accuracy--cost tradeoff.

We first define acquisition cost abstractly, allowing it to capture monetary cost, time, patient burden, risk, or combinations thereof. Let $c_s \in \mathbb{R}_+^{d_s}$ denote the one-time costs of the contextual descriptors, and $c_x \in \mathbb{R}_+^{d}$ denote the per-timepoint costs of the temporal features. For an acquisition trajectory $\{m_t\}_{t=1}^T$, the total cost is
\begin{equation}
\label{eq:total_cost}
\mathrm{Cost}(m_s,\{m_t\}_{t=1}^T)
=
\underbrace{c_s^\top m_s}_{\text{contextual}}
+
\underbrace{\sum_{t=1}^{T} c_x^\top m_t}_{\text{temporal}}.
\end{equation}
Using the predictor in Eq.~\eqref{eq:pred}, we obtain $\hat{y}_t$ from the accrued history up to time $t$. This yields the following accuracy--cost tradeoff objective:
\begin{equation}
\label{eq:objective}
\sum_{t=1}^T
\mathcal{L}_{\mathrm{pred}}(\hat{y}_t, y_t)
+
\lambda\left(
c_s^\top m_s
+
\sum_{t=1}^T c_x^\top m_t
\right),
\end{equation}
where $\lambda > 0$ is an \emph{application-specific} parameter controlling the tradeoff between supervised prediction loss $\mathcal{L}_{\mathrm{pred}}$ (e.g., cross-entropy) and the cost of contextual and temporal acquisitions.

\subsection{The \OurMethod Model}
\label{sec:REACT_components}
\OurMethod is an end-to-end framework that jointly learns a cost-effective subset of onboarding descriptors and personalized temporal acquisition plans under the objective in Eq.~\eqref{eq:objective}. The architecture consists of three interacting modules:
\begin{itemize}[leftmargin=*,topsep=0pt, noitemsep, wide]
    \item \textbf{Context Selector $\alpha \in \mathbb{R}^{d_s}$:} a learned vector $\alpha$ that parameterizes a binary context mask $m_s \in \{0,1\}^{d_s}$ at $t=0$, specifying which onboarding descriptors to acquire for the best downstream accuracy--cost tradeoff.

    \item \textbf{Longitudinal Planner $\pi_\theta$:} a neural network that takes the temporal history up to $t$, the current timestep $t$, and the acquired onboarding context $\tilde{s}$ as input, and outputs logits that parameterize future temporal acquisition masks $m_{t'} \in \{0,1\}^d$ for $t' > t$. 

    \item \textbf{Predictor $f_\phi$:} a classification network as defined in \autoref{sec:formulation}. For a target timestep $t'\geq t$, it maps the acquired context $\tilde{s}$ and temporal history $\mathcal{H}_t$ to class probabilities $\hat{y}_{t'} \in \mathbb{R}^C$.
\end{itemize}

\paragraph{\textbf{The Challenge of Discrete Decisions}} While this architecture intuitively models the two-phase acquisition process, jointly training the parameters $\alpha$, $\theta$, and $\phi$ presents a fundamental optimization challenge. The target masks $m_s$ and $m_t$ are discrete variables (i.e., binary masks). These discrete variables are not differentiable, making it impossible to route gradients from the downstream prediction loss and acquisition cost back to the planner $\pi_\theta$ and context selector $\alpha$ using standard backpropagation. Therefore, to enable end-to-end training of this framework, we employ Gumbel-Sigmoid relaxation~\citep{Maddison2016TheCD} with a straight-through estimator (Sec.~\ref{sec:gumble_sigmoid}).

\subsection{Differentiable Optimization via Relaxation}
\label{sec:gumble_sigmoid}

To handle discrete acquisition decisions, \OurMethod replaces direct binary sampling with the Gumbel-Sigmoid relaxation \cite{Jang2016CategoricalRW, Maddison2016TheCD}, enabling gradients from the prediction loss and acquisition cost to flow through the acquisition masks.

For any logit $l_j$ produced by $\alpha$ or $\pi_\theta$, we inject stochasticity by sampling Gumbel noise
\begin{equation}
g_j = -\log\bigl(-\log u_j\bigr), \qquad u_j \sim \mathrm{Uniform}(0,1).
\end{equation}
Given temperature $\tau > 0$, we then compute a continuous relaxation $\tilde{m}_j \in (0,1)$ as
\begin{equation}
\tilde{m}_j = \sigma\!\left(\frac{l_j + g_j}{\tau}\right).
\end{equation}
During the forward pass, we discretize via thresholding,
\begin{equation}
\hat{m}_j = \mathbbm{I}[\tilde{m}_j > 0.5].
\end{equation}
To retain differentiability during backpropagation, we use the straight-through estimator
\begin{equation}
\label{eq:stop_grad}
\mathcal{G}(l_j) = \tilde{m}_j + \mathrm{sg}(\hat{m}_j - \tilde{m}_j),
\end{equation}
where $\mathrm{sg}(\cdot)$ denotes the stop-gradient operator, so $\hat{m}_j - \tilde{m}_j$ is treated as constant during backpropagation. 
Thus, $\mathcal{G}(l_j)=\hat{m}_j$ in the forward pass, while gradients are taken with respect to the continuous proxy $\tilde{m}_j$ in the backward pass.

Applying $\mathcal{G}(\cdot)$ element-wise to the $d_s$ contextual logits yields the binary context mask $\hat{m}_s \in \{0,1\}^{d_s}$, and applying it to the $d$ temporal logits at timestep $t$ yields the binary temporal mask $\hat{m}_t \in \{0,1\}^{d}$.

\begin{figure}[t]
  \centering
  \includegraphics[width=.90\linewidth, trim={0 0.0cm 0cm 0}, clip]{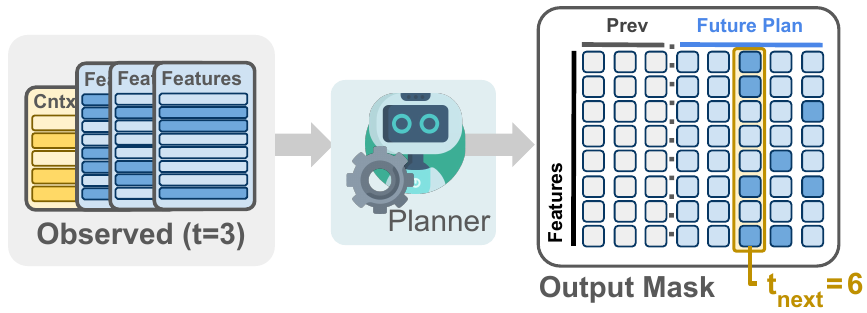}
  \caption{\textbf{Longitudinal planner.} Given the onboarding context and longitudinal measurements observed up to time $t$, the planner outputs a binary acquisition mask over future feature--time pairs. The left portion of the mask corresponds to previously acquired measurements (unused), while the right portion specifies the future acquisition plan, judged against the cost-benefit objective. The earliest selected future time defines the next acquisition time, $t_{\mathrm{next}}$.}
  \label{fig:planner}
\end{figure}

\paragraph{\textbf{Differentiable Loss}}
We train the context selector and planner (see Fig.~\ref{fig:planner}) directly to produce future acquisition plans that optimize the accuracy--cost objective in Eq.~\eqref{eq:objective}. Consider an instance with temporal features $x \in \mathbb{R}^{T \times d}$, contextual descriptors $s \in \mathbb{R}^{d_s}$, labels $y \in \{1,\ldots,C\}^{T}$, and a given temporal mask $\mathbf{M}^{\mathrm{prev}}_{t} \in \{0,1\}^{t \times d}$ representing an observation state at time $t$.

For notation, let $\texttt{EX}_T(v)$ denote zero-padding of $v \in \mathbb{R}^{t \times d}$ to length $T$, let $\texttt{K}_{>t}(v)$ set entries at times $1,\dots,t$ to zero while keeping values after $t$, and let $\texttt{K}_{\leq t}(v)$ set entries at times $t+1,\dots,T$ to zero while keeping values up to and including $t$. We then define the relaxed future acquisition plan
$$
P_{>t}
=
\texttt{K}_{> t}\!\left(
\mathcal{G}\big(
\pi_\theta(\texttt{EX}_T(\mathbf{M}^\mathrm{prev}_{t}) \odot x,\; t,\; \tilde{s})
\big)
\right),
$$
the induced mask available up to time $t' > t$ as
$$
M_{\leq t'}
=
\texttt{EX}_T(\mathbf{M}^\mathrm{prev}_{t})
+
\texttt{K}_{\leq t'}(P_{>t}),
$$
and the relaxed context mask as $\hat{m}_s = \mathcal{G}(\alpha)$.

The total relaxed loss for planning after time $t$ is
\begin{equation}
\label{eq:loss_total}
\begin{aligned}
&\mathcal{L}_\mathrm{REACT}(\alpha, \theta, \phi;\, x, s, y, t, \mathbf{M}^\mathrm{prev}_{t})\\ &=
\underbrace{\sum_{t'=t+1}^{T} \mathcal{L}_{\text{pred}}\!\Big(
\overbrace{f_\phi\!\left(M_{\leq t'} \odot x,\, \hat{m}_s \odot s,\, t'\right)}^{\text{Prediction at } t'},
\, y_{t'}\Big)}_{\text{Prediction Loss for Plan}} \\ 
& \quad + \underbrace{\lambda\,\mathbf{1}_T^\top \left(P_{>t}\right) c_x}_{\text{Temporal Acquisition Cost}}
+ \underbrace{\lambda\, c_s^\top \hat{m}_s.}_{\text{Context Acquisition Cost}}
\end{aligned}
\end{equation}

This relaxed objective enables stable joint gradient updates of $\alpha$, $\theta$, and $\phi$. Notably, $\alpha$ is trained not only to improve prediction directly, but also to select onboarding context that supports better future acquisition decisions by the planner $\pi_\theta$. Moreover, Eq.~\eqref{eq:loss_total} corresponds to a relaxed form of the cost--benefit objective in Eq.~\eqref{eq:objective}.
Below, we describe the training procedure, which uses mini-batches and dynamic rollouts of the current planner to generate observation states for minimizing Eq.~\eqref{eq:loss_total}.

\subsection{Training Procedure}
\label{sec:training}

\begin{algorithm}[ht]
\caption{Self-Iterative Training of \OurMethod}
\label{alg:training}
\begin{algorithmic}[1]
    \REQUIRE Training data $\mathcal{D}_{\text{train}}$, planner $\pi_\theta$, context logits $\alpha$, 
    predictor $f_\phi$, learning rate $\eta$.
    
    \FOR{iteration $i = 1$ to $N$}
        \STATE Sample batch $\mathcal{B} \sim \mathcal{D}_{\text{train}}$
        
        \STATE \textcolor{blue}{\textbf{// Step 1: Forward pass to roll out current policy to collect on-policy states}}
        \FOR{each $(x, s, y) \in \mathcal{B}$}
            \STATE $\hat{m}_s \leftarrow \mathcal{G}(\alpha)$;\quad 
            $\tilde{s} \leftarrow \hat{m}_s \odot s$
            \STATE Initialize $\hat{\mathcal{H}}_0 \leftarrow \mathbf{0} \in \mathbb{R}^{T \times d}$,\  $\mathbf{M}^\mathrm{prev}_{0} \leftarrow \mathbf{0} \in \mathbb{R}^{T \times d}$
            \FOR{$t = 1$ to $T$}
            
                \STATE $\hat{m}_t {\leftarrow} \mathcal{G}(\pi_\theta(\hat{\mathcal{H}}_{t-1};\, t;\, \tilde{s}))[t]$ 
                \textcolor{blue}{\textbf{// query planner output for } $t$ \textbf{based on past history}}
                \STATE $\hat{x}_t \leftarrow \hat{m}_t \odot x_t$
                \STATE $\hat{\mathcal{H}}_t \leftarrow \hat{\mathcal{H}}_{t-1}$;\quad 
                $\hat{\mathcal{H}}_t[t] \leftarrow \hat{x}_t$
                \textcolor{blue}{\textbf{// update history}}
                \STATE $\mathbf{M}^\mathrm{prev}_t \leftarrow \mathbf{M}^\mathrm{prev}_{t-1}$;\quad 
                $\mathbf{M}^\mathrm{prev}_t[t] \leftarrow \hat{m}_t$
                \textcolor{blue}{\textbf{// update history mask}}
            \ENDFOR
            \STATE Store trajectory $\tau = \{(x, s, y, t, \mathbf{M}^\mathrm{prev}_{t})\}_{t=0}^T$ in $\mathcal{B}_{\text{roll}}$
        \ENDFOR

        \STATE \textcolor{blue}{\textbf{// Step 2: Compute loss over on-policy trajectories and the ground truth label y }}
        \STATE $\mathcal{L} = \dfrac{1}{|\mathcal{B}_{\text{roll}}|} 
        \displaystyle\sum_{(x,s,y,t, \mathbf{M}^\mathrm{prev}_{t}) \in \mathcal{B}_{\text{roll}}} 
        \hspace{-1.5em}\mathcal{L}_\mathrm{REACT}(\alpha, \theta, \phi;\, x, s, y, t, \mathbf{M}^\mathrm{prev}_{t})$
        \STATE \textcolor{blue}{\textbf{// Step 3: Update w/ gradient descent (or \emph{other gradient based optimizer})}}
        \STATE $\alpha \leftarrow \alpha - \eta\nabla_{\alpha}\mathcal{L}$;\quad 
        $\theta \leftarrow \theta - \eta\nabla_\theta\mathcal{L}$;\quad 
        $\phi \leftarrow \phi - \eta\nabla_\phi\mathcal{L}$
    \ENDFOR
\end{algorithmic}
\end{algorithm}

\paragraph{\textbf{Self-Iterative Training}} 
The planner is trained via \emph{self-iterative training}, where the current planner rolls out on-policy trajectories that are then reused as training states.
This allows the planner to improve beyond the offline reference plans by training on states it induces itself---similar in spirit to DAgger-style iterative imitation learning~\cite{ross2011reduction}, but driven by direct gradient optimization with ground-truth labels rather than imitation targets.
Training alternates between the following steps (details in Algorithm~\autoref{alg:training}): (1) \textbf{On-policy rollout:} sample a batch $\mathcal{B}$ from $\mathcal{D}_{\text{train}}$, and execute the current planner $\pi_\theta$ and context selector $\alpha$ to collect  trajectories; (2) \textbf{Joint gradient update:} compute $\mathcal{L}$ from Eq.~\eqref{eq:loss_total} and jointly update $\alpha$, $\theta$, and $\phi$ by backpropagating through the ST-Gumbel-Sigmoid gates. Note that the rollouts in Algorithm \autoref{alg:training} may contain skipped timepoints or early termination when the planner outputs zero-masks (\texttt{line 8}).

\subsection{Inference-Time Acquisition Algorithm}
\label{sec:inference}

At inference time, \OurMethod operates sequentially as summarized in Algorithm~\ref{alg:react_infer}. Gumbel noise is removed, and all acquisition masks are obtained deterministically by hard-thresholding the learned logits.\looseness-1

The process begins at onboarding ($t=0$), where the learned context mask
$m_s = \mathbb{I}[\sigma(\alpha) > 0.5]$
determines which contextual features to acquire. The adaptive longitudinal phase then proceeds over $t \in \{1,\dots, T\}$. At each step, the planner $\pi_\theta$ evaluates the current history $\mathcal{H}_{t}$ together with the acquired context $\tilde{s}$ to produce a future temporal acquisition plan $M$. \OurMethod acquires features only at the next selected timepoint, skipping acquisition at intermediate timesteps, and may terminate early if no further acquisitions are planned. Predictions $\hat{y}_t$ are made at each timestep using the information available up to $t$, and the process continues until termination or the horizon $T$ is reached.





\begin{algorithm}[ht]
\caption{\OurMethod Inference-Time Acquisition}
\label{alg:react_infer}
\begin{algorithmic}[1]
\REQUIRE trained context selector $\alpha$; trained planner $\pi_\theta$; predictor $f_\phi$.
\ENSURE Predictions $\{\hat{y}_t\}_{t=1}^T$

\STATE Initialize: $\texttt{predictions} \leftarrow [\,]$
\STATE Initialize acquired history: $\mathcal{H} \leftarrow (\mathbf{0}, \dots, \mathbf{0}) \in \mathbb{R}^{T \times d}$

\STATE \textcolor{blue}{\textbf{// Stage 1: Contextual descriptor acquisition}}
\STATE $m_s \leftarrow \mathbb{I}[\sigma(\alpha) > 0.5]$ 
\STATE $\tilde{s} \leftarrow \mathbf{0}$; \ \text{Acquire } $\mathbf{s}_j$ \text{for} $m_{s,j}>0$
\STATE $\mathbf{M} \leftarrow \mathbb{I}\big[\sigma\big(\pi_\theta(\mathcal{H},\, 0,\, \tilde{s})\big) > 0.5\big]$
\STATE $t^\mathrm{next} \leftarrow \min \{t' \mid \|\mathbf{M}[t']\|>0 \}$ \textcolor{blue}{\textbf{// next time with nonzero mask}}
\STATE $m^\mathrm{next} \leftarrow \mathbf{M}[t^\mathrm{next}]$ \textcolor{blue}{\textbf{// next acquisition mask}}

\STATE \textcolor{blue}{\textbf{// Stage 2: Adaptive temporal acquisition and prediction}}
\FOR{$t = 1$ \text{ to } $T$}

    \IF{$t==t^\mathrm{next}$}        
        \STATE \textcolor{blue}{\textbf{// Acquire and replan}}
        \STATE $\mathbf{x} \leftarrow \mathbf{0}$
        \STATE \text{Acquire } $\mathbf{x}_j$ \text{for} $m^\mathrm{next}_j>0$
        \STATE $\mathcal{H}[t] \leftarrow \mathbf{x}$
        \STATE $\mathbf{M} \leftarrow \texttt{K}_{>t} \Big[\mathbb{I}\big[\sigma\big(\pi_\theta(\mathcal{H},\, t,\, \tilde{s})\big) > 0.5\big]\Big]$ \textcolor{blue}{\textbf{// future mask plan}}
        \STATE $t^\mathrm{next} \leftarrow \min \{t' \mid \|\mathbf{M}[t']\|>0 \}$ \textcolor{blue}{\textbf{// or $\varnothing$ for termination}}
        \STATE $m^\mathrm{next} \leftarrow \mathbf{M}[t^\mathrm{next}] $ 
    \ENDIF

    \STATE \textcolor{blue}{\textbf{// Predict}}

    \STATE $\hat{y}_t \leftarrow f_\phi(\mathcal{H},\, \tilde{s},t)$
    \STATE append $\hat{y}_t$ to \texttt{predictions}

\ENDFOR

\STATE \textbf{Return} \texttt{predictions}
\end{algorithmic}
\end{algorithm}

\section{Experiment}


\subsection{Datasets}
We evaluate \OurMethod on four real-world longitudinal datasets spanning two application families: behavioral longitudinal datasets (CHEEARS, ILIADD) and clinical longitudinal datasets (OAI, ADNI).
Full dataset and feature details can be found in Appx.~\ref{appendix:dataset}.

\subsubsection{Behavioral Longitudinal Datasets}
\paragraph{\textbf{CHEEARS}} The CHEEARS dataset \cite{Ringwald2025CommonAU} comprises ecological momentary assessment (EMA) data from 204 college students. Each participant completed daily surveys assessing affect, drinking behaviors, and social context. For this study, we divided the data into sliding windows of  10 consecutive days and set the target to be next-day drinking behavior prediction (binary classification). We split the data temporally by a fixed cutoff date, such that training instances correspond to observations before the cutoff and test instances correspond to observations after it. The onboarding data contains demographics, AUDIT alcohol screening scores, drinking motives (DMQ: social, coping, enhancement, conformity), alcohol consequences (YAACQ: 9 subscales), personality traits (NEO Big Five), and interpersonal functioning (IIP: dominance, affiliation, elevation). The temporal data contains daily affect (10 items: happy, stressed, anxious, etc.), drinking urges/plans/quantity expectations, and social experiences. We assigned an equal feature cost of 1 to all features. The dataset is split into train/val/test ($60.6\%$, $18.9\%$, $20.5\%$).

\paragraph{\textbf{ILIADD}}
The ILIADD (Intensive Longitudinal Investigation of Alternative Diagnostic Dimensions) \cite{Ringwald2025DoYF} dataset is a fully remote ambulatory assessment study ($N=544$ participants with $\geq 5$ EMA surveys completed) for mental health treatment history (81\% 
reporting prior treatment). The contextual features consist of demographics and scales derived from the HiTOP-SR, factor scales, symptom scales, and alcohol use scales. In the study,  this set of questions was prompted to participants 8 times per day. Therefore, we used these collected answers throughout the day as temporal features; these assessed momentary positive affect,  energy, stress, and impulsivity (4 items). We split the data temporally by a fixed cutoff date/time. For this study, we divided the data into consecutive sliding windows of 10 time steps, and the label negative affect is predicted for each timepoint. We assigned an equal feature cost of 1 to all features. The dataset is split into train/val/test ($69.5\%$, $15.2\%$, $15.2\%$).

\subsubsection{Clinical Longitudinal Datasets}
\paragraph{\textbf{OAI}} The Osteoarthritis Initiative (OAI) \footnote{\url{https://nda.nih.gov/oai/}} is a longitudinal cohort that tracks knee osteoarthritis progression with annual follow-up up to $96$ months for $4{,}796$ patients using imaging and clinical assessments. We select $d_s=10$ contextual features (e.g., sex, race, age) and $d=17$ temporal features (including clinical measurements and $10$ joint space width - JSW - features extracted from knee radiography) over $T=7$ visits.
Following \citet{chen2024unified, nguyen2024active}, we consider two prediction targets at each visit: (i) Kellgren--Lawrence grade (KLG \cite{kellgren1957radiological}; range $0\!-\!4$), where we merge grades 0 and 1, and (ii) WOMAC pain \cite{mcconnell2001western} (range $0\!-\!20$), where we define $\mathrm{WOMAC}<5$ as no pain and $\ge 5$ as pain.
We assigned lower costs to low-effort questionnaire/clinical variables (e.g., $0.3$-$0.5$) and higher costs to JSW extracted from knee radiograph (e.g., $0.8$-$1.0$). Following \citet{chen2024unified}, we split the dataset into train/val/test ($50\%$, $12.5\%$, $37.5\%$).

\paragraph{\textbf{ADNI}} The Alzheimer’s Disease Neuroimaging Initiative (ADNI)\footnote{\url{https://adni.loni.usc.edu/}} \cite{petersen2010alzheimer} is a longitudinal, multi-center, observational study for tracking Alzheimer’s disease progression. Following \cite{qin2024risk}, we use a benchmark of $N=1{,}002$ participants with regular follow-up every six months and consider the first $T=12$ visits. The onboarding context consists of $d_s=7$ descriptors (e.g., age, gender, Functional Activities Questionnaire), while the temporal feature bank contains $d=4$ imaging biomarkers: FDG and AV45 from PET, and Hippocampus and Entorhinal from MRI. At each visit, the goal is to predict the patient’s current disease status as normal cognition, mild cognitive impairment, or Alzheimer’s disease \cite{o2008staging, o2010validation}. Context acquisition costs are set to $0.3$, and higher costs are assigned to PET biomarkers ($1.0$ each) and MRI biomarkers ($0.5$ each) to reflect the greater expense and burden of PET and MRI imaging. Following \citet{qin2024risk}, we split the dataset into train/val/test ($64\%$, $16\%$, $20\%$).

\subsection{Baselines}
We compare our framework against several strong AFA baselines, focusing on RL-based methods for longitudinal settings. These include \textbf{ASAC}~\cite{yoon2019asac}, an actor-critic approach that jointly optimizes feature selection and prediction; \textbf{RAS}~\cite{qin2024risk}, which dynamically determines acquisition timing and feature subsets over continuous time; and its variant \textbf{AS}~\cite{qin2024risk}, which enforces uniform acquisition intervals.\looseness-1

We also include \textbf{DIME}~\cite{gadgil2023estimating}, a representative non-longitudinal AFA framework. To adapt it to our setting, we restrict its action space to the current and future timesteps only. This gives DIME an advantage, as it may acquire multiple features at the present timestep before advancing.


Importantly, existing baselines do not natively distinguish the onboarding context $s$ from the temporal measurements $x_t$. To evaluate them fairly, we modify their observation space at every timestep $t$ to include static features as $x'_{t} = [s, x_t] \in \mathbb{R}^{d_s + d}$. Thus, the baselines can acquire the context at any time if they missed it earlier, giving them an advantage. However, the baselines must learn to acquire context early and avoid redundantly spending the acquisition cost $c_s$ for the same unchanging features at future steps. To ensure a fair comparison, we started from the authors’ official implementations and applied the above adaptations. We tuned the hyperparameters for each method on the validation set and report the resulting test performance; exhaustive setup details are in Appx.~\ref{appendix:baseline}.

\begin{figure*}[t]
\centering
\includegraphics[width=0.95\linewidth]{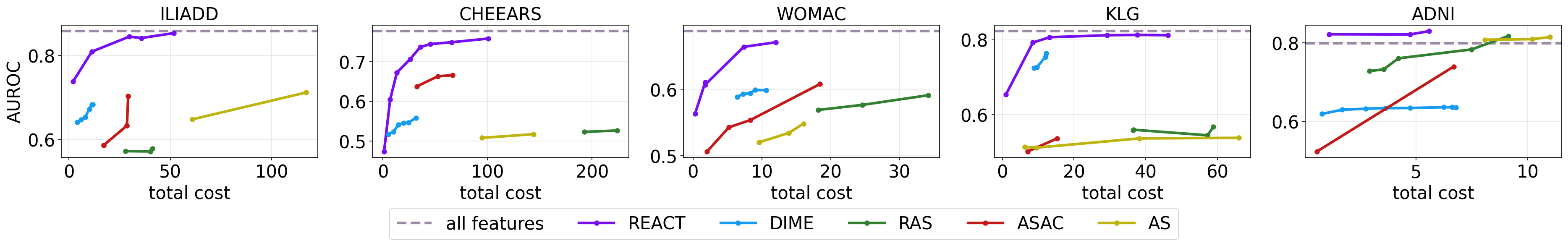}
\caption{AUROC/total cost of models 
  across various average acquisition costs (budgets) on test data. The dashed line is evaluating the pretrained classifier from REACT with all features available.}
\label{fig:performance_cost}
\end{figure*}

\begin{figure}[t]
  \centering
  \includegraphics[width=\linewidth, trim={0 0 11cm 0}, clip]{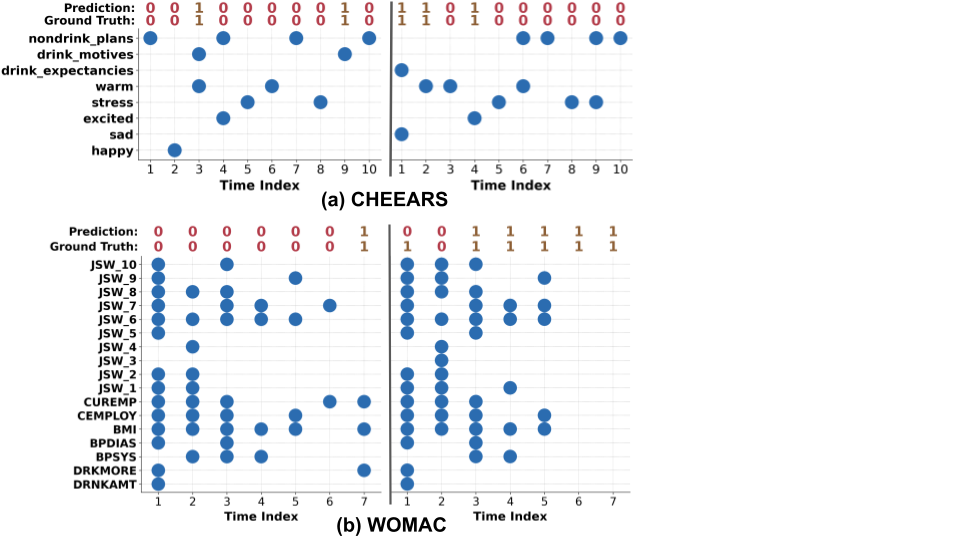}
  \caption{Example feature acquisition rollouts using \OurMethod for two distinct instances from the (a) CHEEARS and (b) WOMAC test sets. For visual clarity, only the selected features for these instances are displayed.}
  \label{fig:instance_rollout}
\end{figure}
\subsection{Implementation Details}

\subsubsection{Predictor Architecture \& Training}For prediction at target timestep $t'$, the predictor takes as input the acquired temporal history $\mathcal{H}_{t'}$, acquired context $\tilde{\mathbf{s}}$, and normalized time indicator for $t'$. These are passed through a 3-layer MLP producing class logits (see details in Appx.~\ref{appendix:experiment}). The predictor is pre-trained with random masking, then jointly trained with the planner during acquisition policy learning.


\subsubsection{Planner Architecture \& Training}
The planner $\pi_\theta$ consists of 3 hidden layers with dimensions [512, 256, 128] and ReLU activations. The planner maps the acquired data to a mask over future acquisitions (see Sec.~\ref{sec:REACT_components}). We trained the planner network with a total of 1K batches utilizing our self-iterative training procedure (Sec. \ref{sec:training}). Additional details may be found in Appx.~\ref{appendix:experiment}. 

Please see timing results in Appx. Fig.~\ref{fig:runtime_comparison}, where \OurMethod achieves training and inference times that are faster than or comparable to recent deep learning approaches, such as DIME, RAS, and AS. We will open-source the \OurMethod code upon publication. 
\subsection{Performance-Cost Tradeoff} 
In Fig.~\ref{fig:performance_cost}, we evaluate \OurMethod on five longitudinal prediction tasks and plot the AUROC vs.~the total average acquisition cost  (see AURPC figures, which mostly follow the same trends, in Appx.~\ref{appendix:add_results}). Here, the total cost includes both the one-time onboarding context costs and temporal feature acquisition costs. For \OurMethod, one can learn the policy at different cost budget ranges by sweeping the cost coefficient $\lambda$ in Eq.~\eqref{eq:objective}, which controls the tradeoff between prediction loss and acquisition cost (see Appx. Tab.~\ref{tab:react_costs} for the values we use for $\lambda$). For the baselines, we analogously vary each method's cost-sensitive hyperparameter to obtain different average costs (more details in Appx.~\ref{appendix:baseline}). Thus, each data point in the curve corresponds to a different performance-cost setting, and a method is preferred if it achieves higher performance at a lower total cost (higher toward the upper left). That is, y-axes show the predictive performance (e.g., AUROC or AUPRC) of the models (at various cost/benefit tradeoffs) over the test set. Correspondingly, we record the average cost incurred during inference (at various cost/benefit tradeoffs) from acquisitions.

Since the total acquisition costs include the one-time onboarding context cost, part of REACT's advantage comes from learning a selective subset of contextual descriptors rather than acquiring all onboarding variables. This learned context selector $\alpha$ provides informative initialization for the downstream temporal planner $\pi_\theta$; examples of dataset-level context selection are provided in Appx.~\ref{appendix:onboarding_context}, and feature descriptions are given in Appx.~\ref{appendix:dataset}.
\looseness - 1

\begin{figure*}[h]
  \centering

  \begin{minipage}[t]{0.32\textwidth}
    \centering
    \includegraphics[width=\linewidth]{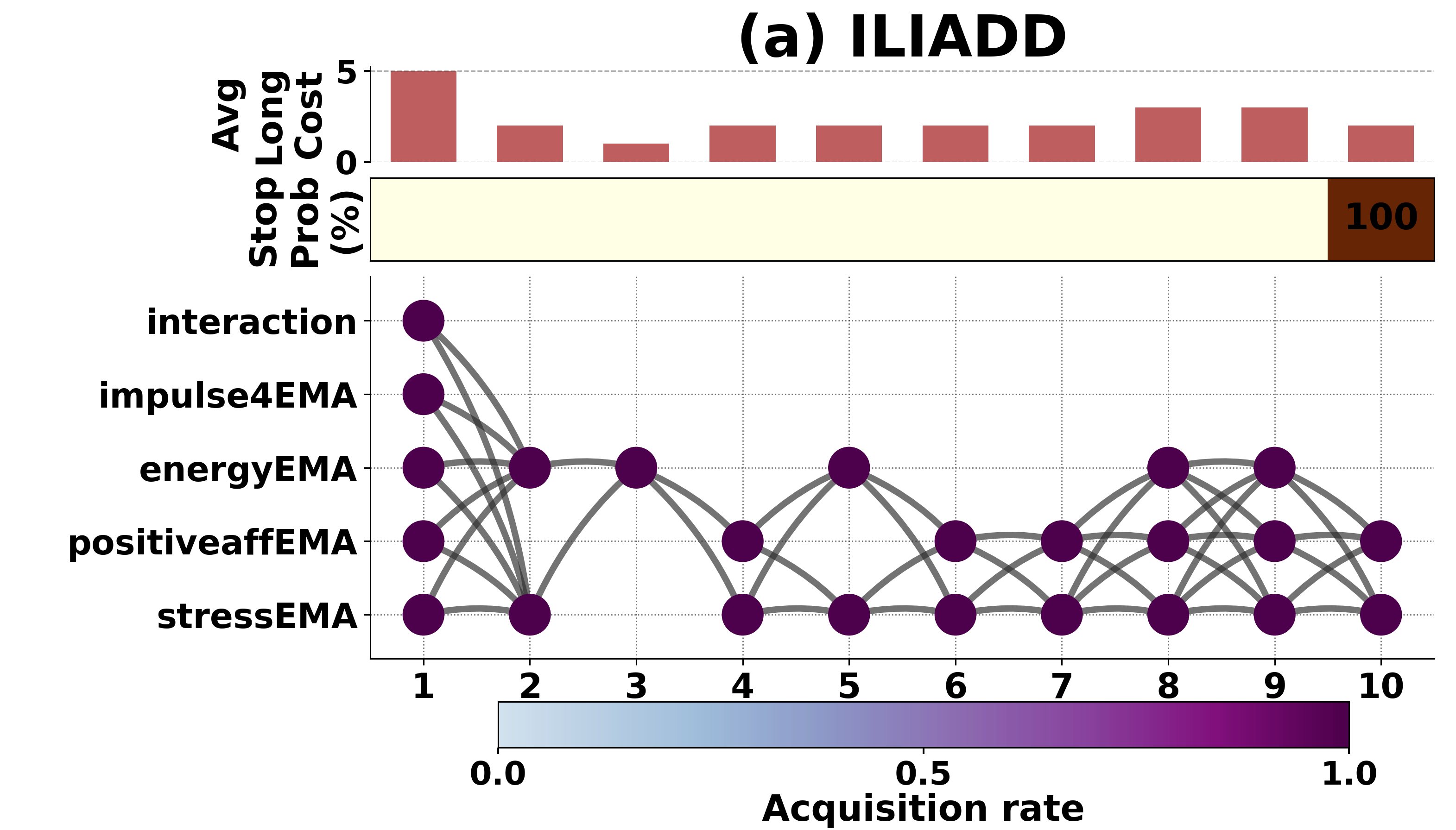}
  \end{minipage}\hfill
  \begin{minipage}[t]{0.32\textwidth}
    \centering
    \includegraphics[width=\linewidth]{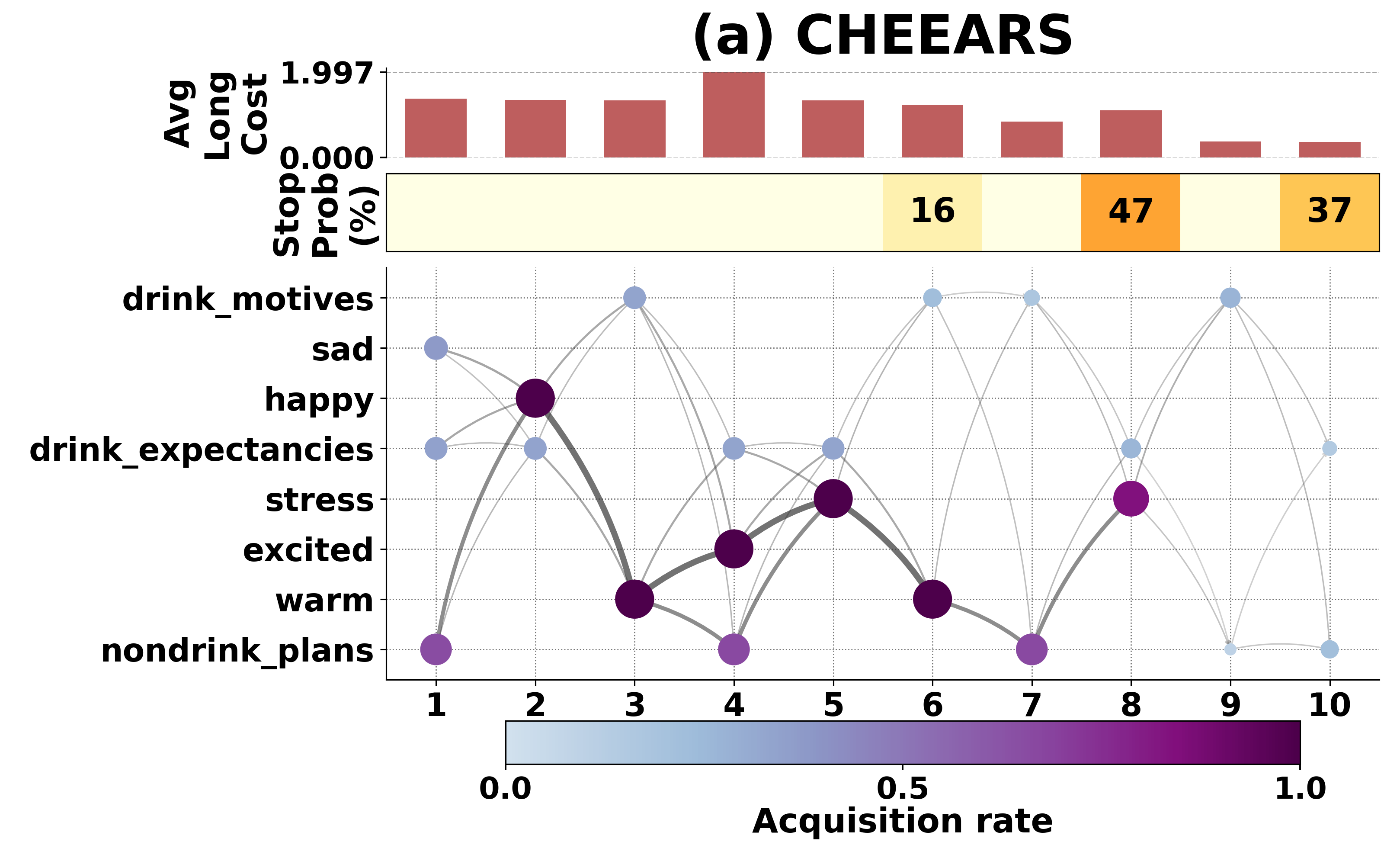}
  \end{minipage}\hfill
  \begin{minipage}[t]{0.32\textwidth}
    \centering
    \includegraphics[width=\linewidth]{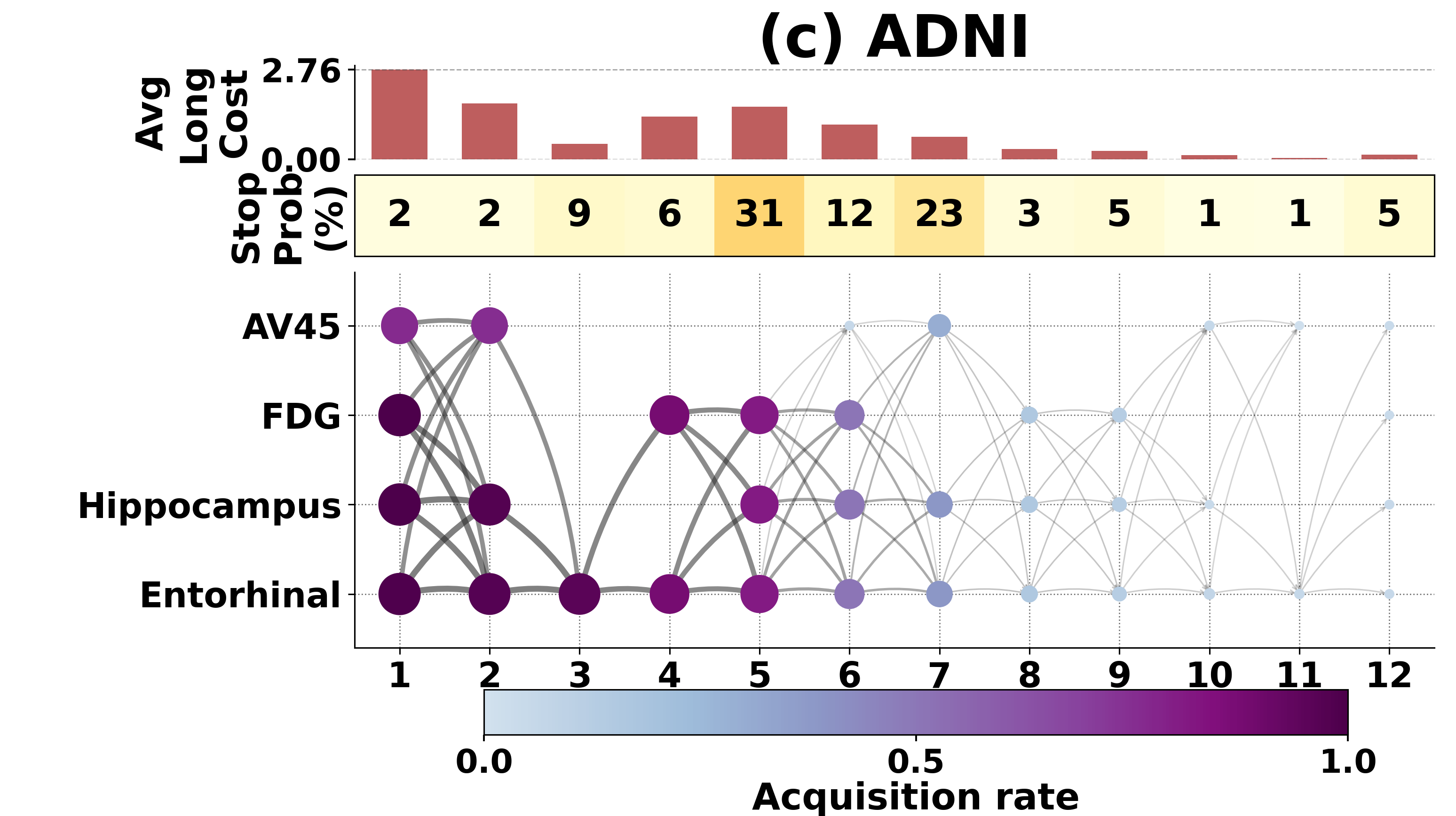}
  \end{minipage}

  \vspace{0.6em}

  \begin{minipage}{0.68\textwidth}
    \centering
    \begin{minipage}[t]{0.47\linewidth}
      \centering
      \includegraphics[width=\linewidth]{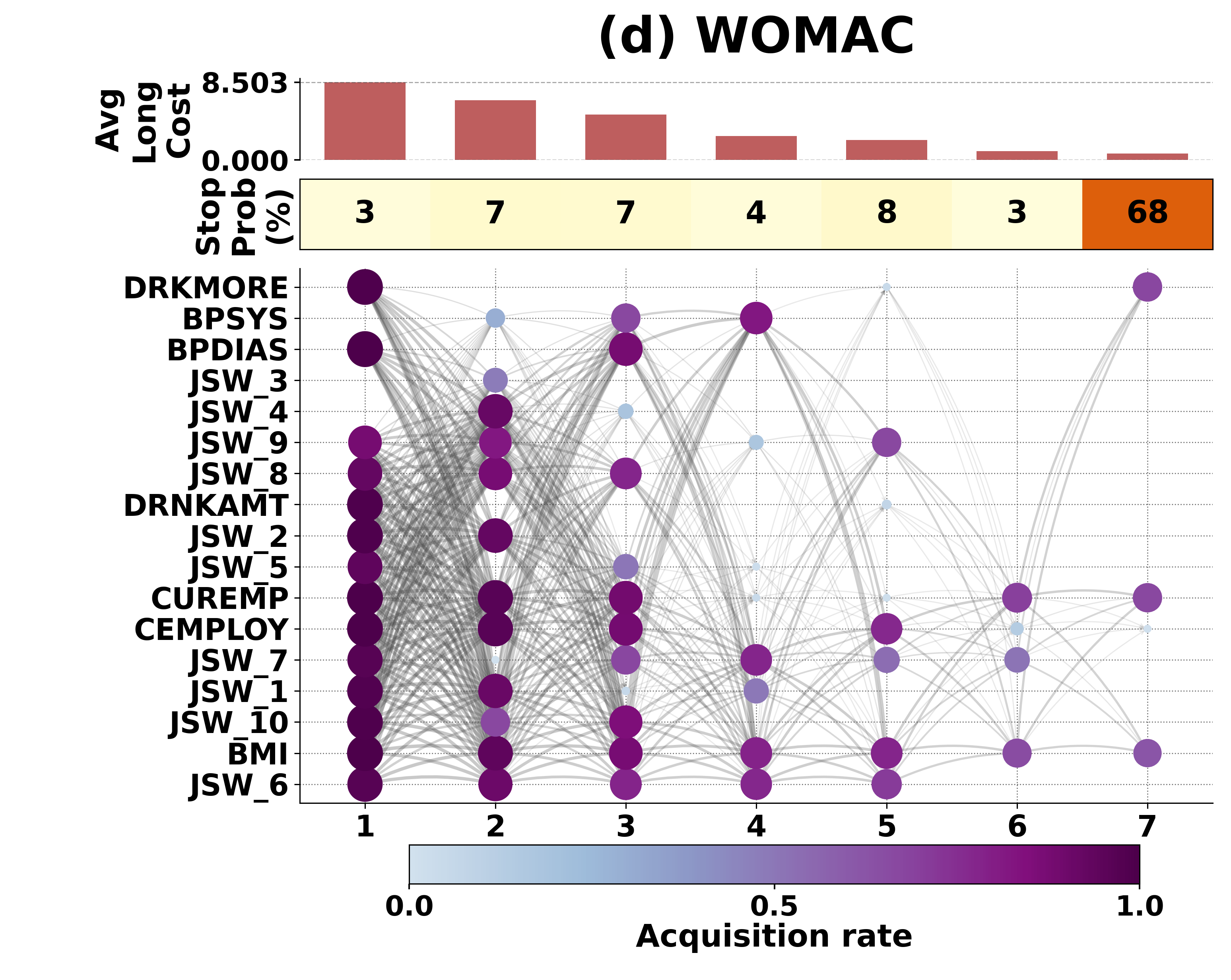}
    \end{minipage}\hfill
    \begin{minipage}[t]{0.47\linewidth}
      \centering
      \includegraphics[width=\linewidth]{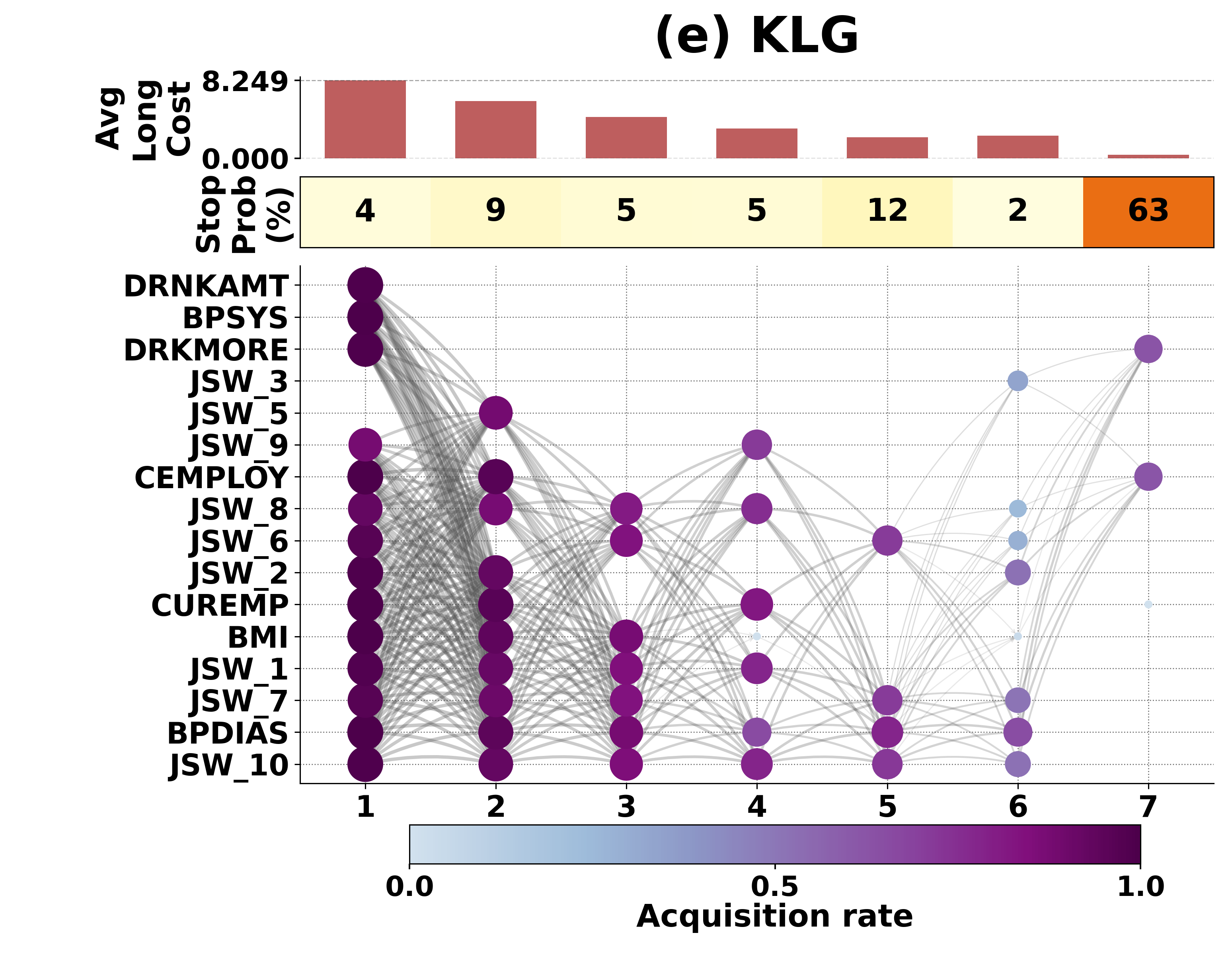}
    \end{minipage}
  \end{minipage}

  \caption{Qualitative visualization of longitudinal acquisition dynamics by \OurMethod across datasets. Metrics are reported per dataset on the test set as (Total/Longitudinal Costs $\mid$ AUROC/AUPRC): ILIADD (35.739/23.993 $\mid$ 0.842/0.706),  CHEEARS (13.275/11.275 $\mid$ 0.673/0.540), ADNI (12.635/10.535 $\mid$ 0.823/0.678), WOMAC (30.217/28.141 $\mid$ 0.670/0.355), KLG (29.243/26.869 $\mid$ 0.812/0.621). For visual clarity, only the selected features are displayed, and the stop probabilities are rounded.}
  \label{fig:traj_all}
\end{figure*}

\begin{figure}[t]
  \centering
  \includegraphics[width=\linewidth, trim={0 7.4cm 0cm 0}, clip]{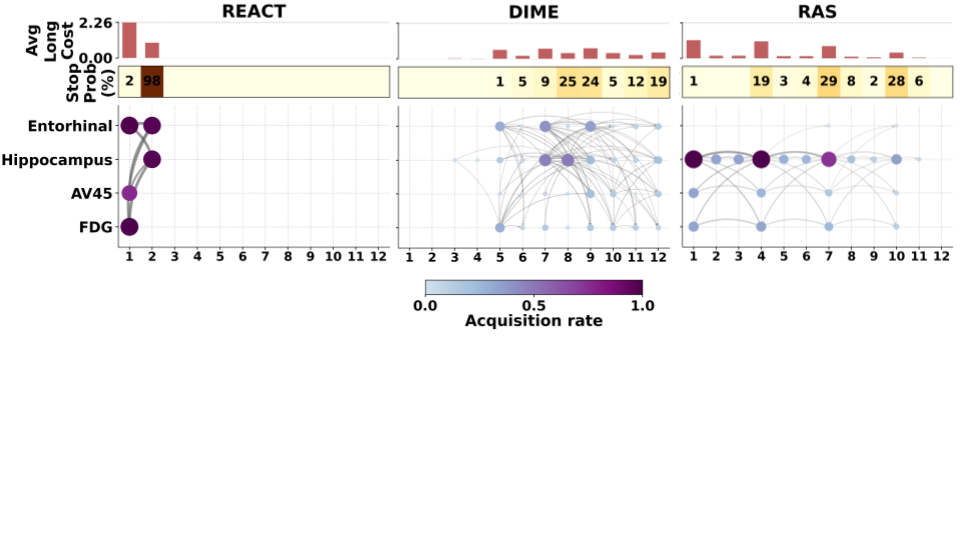}
  \caption{Qualitative comparison of acquisition of \OurMethod, DIME, and RAS on the ADNI dataset. Metrics are reported per framework on the test set as (Total/Longitudinal Costs $\mid$ AUROC/AUPRC): \OurMethod (5.335/3.235 $\mid$ 0.824/0.683), DIME (5.619/3.630 $\mid$ 0.644/0.483), RAS (8.675/4.193 $\mid$ 0.817/0.684)}
  \label{fig:baselines}
\end{figure}


\paragraph{\textbf{Behavioral Datasets.}} On CHEEARS and ILIADD, \OurMethod consistently provides the best overall trade-off. The gains are particularly clear on ILIADD, where it outperforms DIME, RAS, and ASAC across all budgets, approaching the performance of the all-features baseline at a moderate cost.

\paragraph{\textbf{Clinical Datasets}} On KLG and WOMAC, \OurMethod again yields the strongest performance across all budget constraints compared to the baselines. On ADNI, the performance gap narrows, with DIME also performing strongly; however, \OurMethod remains highly competitive and achieves slightly better overall results.
For additional results on the AUPRC metric, please see Appx.~\ref{appendix:add_results}. With AUPRC, \OurMethod shows a similar trend, outperforming the baselines.





\subsection{Analysis of Learned Acquisition Behavior}


To better understand the acquisition policies learned by \OurMethod, we provide additional analyses across datasets. Before discussing these analyses, we first detail the construction of these visualizations. Fig.~\ref{fig:traj_all} and Fig.~\ref{fig:baselines} show the learned acquisition trajectories across datasets. For each dataset, the \textbf{top panel} displays the average temporal acquisition cost incurred per timestep, while the \textbf{middle panel} shows the termination (i.e., stop acquiring) probability distribution. The \textbf{bottom panel} illustrates the temporal acquisition plan as a directed graph across timesteps (y-axis). Nodes represent specific features acquired at a given timestep, with size and color intensity reflecting the overall acquisition frequency. Edges show the directed transitions between acquired nodes: if a sample acquires feature \texttt{m} at one timestep $t$ and feature \texttt{n} at a future acquisition step $t' > t$, a directed edge $\texttt{m}@t \rightarrow \texttt{n}@t'$ is added. The visual weight of the edge (thickness and darkness) corresponds to the fraction of samples exhibiting that transition. For visual clarity, we only show the features actually acquired across the trajectories. 

Fig.~\ref{fig:traj_all} shows that \OurMethod learns markedly different longitudinal acquisition policies across datasets, rather than applying a uniform sparsification strategy. Across tasks, acquisition is often front-loaded, with higher measurement cost in early steps followed by progressively sparser follow-up, suggesting that early observations are frequently most useful for routing later decisions. The stop-probability distributions further indicate adaptive termination behavior: in CHEEARS and ADNI, stopping mass is spread across intermediate and later steps, consistent with selective early stopping once sufficient evidence is gathered, whereas in ILIADD, WOMAC, and KLG, termination under the policy shown here is concentrated closer to the end of the horizon, indicating that continued monitoring is often preferred at this operating point. The trajectory graphs reinforce this interpretation by showing structured transition patterns, where early acquisitions branch into different later feature sequences rather than repeatedly selecting the same variables in a fixed order. 
Fig.~\ref{fig:instance_rollout} further shows that \OurMethod adapts feature acquisition at the instance level: (a) For CHEEARS, only the first instance acquires the \texttt{drink\_expectancies} and \texttt{happy} features, and (b) for WOMAC, only the second instance acquires \texttt{JSW\_3}, and the policy stops acquiring once predictions stabilize, when additional measurements are unlikely to change the predicted label to save the budget.
We note that these visualizations correspond to \OurMethod policies learned for a respective single acquisition cost tradeoff parameter, $\lambda$.

\begin{figure*}[t]
    \centering
    \includegraphics[width=0.95\linewidth]{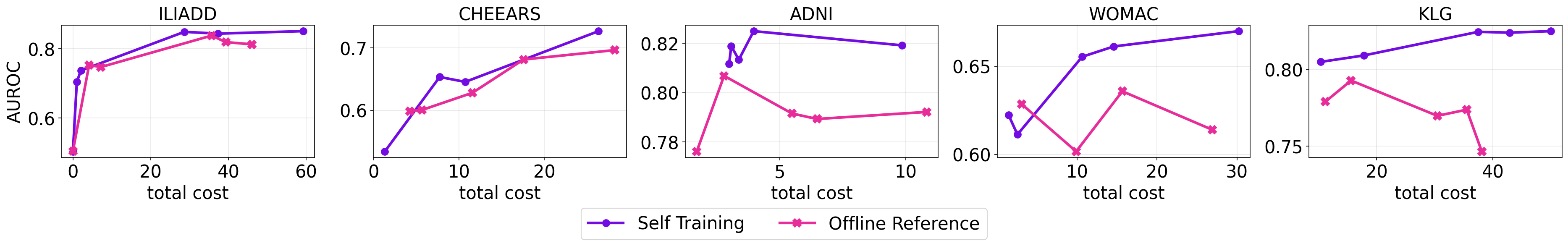}
    \caption{Ablation study of the self-iterative training we used for training \OurMethod (Alg. \ref{alg:training}). We compare AUROC/total cost when planner $\pi_\theta$ trained with (i) self-iterative training from random initialization (Self Training) and (ii) trained using only offline reference states (Offline Reference).}
    \label{fig:ABL_self_train}
\end{figure*}
\begin{figure*}
    \centering
    \includegraphics[width=0.95\linewidth]{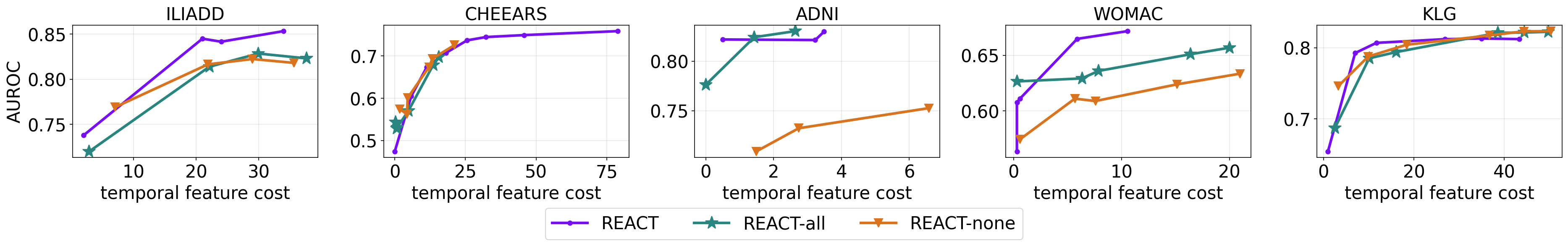}
    \caption{Ablation study on how the learned context selector affects the temporal feature acquisition. AUROC of \OurMethod with learned context descriptors compared to when context descriptors are all acquired (\OurMethod-all) or not at all acquired (\OurMethod-none) across datasets.}
    \label{fig:ABL_context}
\end{figure*}
The dataset-specific patterns are consistent with the underlying measurement types. In CHEEARS, \OurMethod concentrates on a relatively small subset of daily EMA variables, including affective states, drinking-related expectations, and social-context items, with acquisition frequency declining over time. This suggests that for next-day drinking prediction, a limited set of recent mood, motivation, and context measurements often captures much of the useful signal, after which additional daily acquisitions provide diminishing value. In ILIADD, the learned policy repeatedly acquires the compact set of momentary EMA items---positive affect, energy, stress, and impulsivity---across the horizon. Because the target is negative affect at each timepoint and the temporal bank itself consists of repeatedly sampled within-day EMA variables, this sustained reuse of a small dynamic feature set is behaviorally plausible and aligns with the strong predictive performance on this dataset. Notably, this comparatively static acquisition pattern illustrates that \OurMethod adapts the degree of temporal policy dynamicity to the dataset, rather than enforcing diverse trajectories when repeated measurement of the same small feature set is most useful. In ADNI, \OurMethod focuses on the four longitudinal imaging biomarkers---Hippocampus, Entorhinal, FDG, and AV45---with a transition structure suggesting earlier use of lower-cost MRI markers and more selective progression to higher-cost PET biomarkers when needed. The spread of stopping mass across visits is likewise consistent with adaptive escalation when later acquisitions differ substantially in cost and burden. In OAI (WOMAC and KLG), the learned policies (under the choice for $\lambda$) show broader early use of the available longitudinal clinical and radiographic variables, followed by gradual narrowing over later visits. 
These patterns suggest that, when allowed a less restrictive budget, \OurMethod can capitalize on that flexibility by acquiring a wider set of early clinical and radiographic measurements before concentrating on a smaller follow-up subset.
The denser early graphs likely reflect an increased budget of the selected cost--benefit trade-off $\lambda$ rather than an inherent need for broader acquisition. Overall, these visualizations suggest that \OurMethod allocates longitudinal acquisition effort in a task-dependent manner, adapting both feature choice and stopping behavior to the temporal structure, feature semantics, and relative acquisition costs of each dataset.

\paragraph{\textbf{Acquisition Trajectories of \OurMethod vs. Baselines}} 
We compare against DIME and RAS baselines at a similar average cost in Fig.~\ref{fig:baselines}. \OurMethod achieves better performance than DIME at a similar cost and comparable performance to RAS at a lower cost. Moreover, one may see that \OurMethod acquires informative biomarkers early and then terminates quickly ($98 \%$ of trajectories terminate by timestep two). This behavior suggests a time-aware strategy, where the policy invests early to reduce the chance of missing high-value signals, and stops once additional measurements are unlikely to change the prediction. In contrast, DIME and RAS spread the acquisition over time, showing later stopping distribution mass and extended acquisition chains. This suggests the baselines struggle to stop (even) when they have enough information.

\subsection{Ablation of Context Selection}
We ablate the onboarding context selector $\alpha$ against two variants: (i) \OurMethod-all, which acquires all context features, and (ii) \OurMethod-none, which uses none. As shown in Fig.~\ref{fig:ABL_context}, the learned selector achieves higher AUROC at comparable cost on CHEEARS, ILIADD, and WOMAC. On KLG, all three variants perform similarly, suggesting context contributes little predictive value. On ADNI, removing context hurts performance, while \OurMethod remains competitive with \OurMethod-all — indicating the selector learns to acquire useful context without needing all of it. Overall, the benefit comes not from more context but from a learned selective subset that better supports downstream temporal acquisition. AUPRC results in Appx. Fig.~\ref{fig:ABL_contex_auprc} show a consistent pattern.

\subsection{Ablation of Self-Iterative Training}
To validate self-iterative training, we ablate it against a variant trained solely on static offline reference plans (see Appx.~\ref{appendix:warmup}). As shown in Fig.~\ref{fig:ABL_self_train}, self-iterative training consistently outperforms the offline reference across all datasets. The gap is most pronounced on clinical datasets (WOMAC, KLG, ADNI), where the offline variant produces unstable, non-monotonic curves at higher cost budgets — suggesting increasing misalignment between reference plans and the states the planner actually induces at test time. This confirms that exposing the planner to its own rollout states is critical for effective policy learning. Additional AUPRC results are in Appx. Fig.~\ref{fig:ABL_warmup_auprc}.\looseness - 1

\section{Conclusion}
We presented \OurMethod, which formalizes a practical gap in the LAFA literature: the explicit separation of onboarding context from temporal features---a two-phase structure common to real clinical workflows. 
\OurMethod yields an end-to-end differentiable framework for LAFA that jointly optimizes onboarding context selection and adaptive temporal acquisition planning under cost constraints. By replacing discrete optimization with Gumbel-Sigmoid relaxation, \OurMethod trains a unified acquisition policy without the instability and credit-assignment difficulties of RL-based approaches. Importantly, the context selector is optimized not in isolation, but to select onboarding descriptors that best support the downstream planner's cost-benefit tradeoff---learning a population-level context strategy tailored to longitudinal acquisition efficiency.

We provide both quantitative results, such as cost-performance curves, and qualitative analysis that shows feature acquisition processes on five real-world longitudinal tasks with context features spanning behavioral and clinical health domains. Across these datasets, \OurMethod consistently achieves superior predictive performance at the same acquisition cost compared to existing LAFA baselines. Qualitative analysis reveals that \OurMethod learns meaningful, dataset-specific acquisition strategies: acquiring 
informative features early and terminating once predictions stabilize. 
Ablation studies confirm that both the learned context selector and the self-iterative training procedure contribute meaningfully to the superior performance of \OurMethod by identifying a selective subset of onboarding descriptors that best support downstream temporal decisions and exposing the planner to on-policy states that better reflect test-time conditions, respectively. 

We anticipate that this work serves as a foundation for future research on cost-aware longitudinal decision-making. In the longer term, approaches such as \OurMethod may support more efficient, patient-centered care by reducing unnecessary measurement burden, lowering costs, and improving the allocation of expensive diagnostic resources.
\bibliographystyle{ACM-Reference-Format}

\bibliography{bibliography}


\begin{thebibliography}{40}


\ifx \showCODEN    \undefined \def \showCODEN     #1{\unskip}     \fi
\ifx \showISBNx    \undefined \def \showISBNx     #1{\unskip}     \fi
\ifx \showISBNxiii \undefined \def \showISBNxiii  #1{\unskip}     \fi
\ifx \showISSN     \undefined \def \showISSN      #1{\unskip}     \fi
\ifx \showLCCN     \undefined \def \showLCCN      #1{\unskip}     \fi
\ifx \shownote     \undefined \def \shownote      #1{#1}          \fi
\ifx \showarticletitle \undefined \def \showarticletitle #1{#1}   \fi
\ifx \showURL      \undefined \def \showURL       {\relax}        \fi
\providecommand\bibfield[2]{#2}
\providecommand\bibinfo[2]{#2}
\providecommand\natexlab[1]{#1}
\providecommand\showeprint[2][]{arXiv:#2}

\bibitem[Aiyegbusi et~al\mbox{.}({[n.\,d.]})]%
        {aiyegbusikey}
\bibfield{author}{\bibinfo{person}{OL Aiyegbusi}, \bibinfo{person}{J
  Roydhouse}, \bibinfo{person}{SC Rivera}, \bibinfo{person}{P Kamudoni},
  \bibinfo{person}{P Schache}, \bibinfo{person}{R Wilson}, \bibinfo{person}{R
  Stephens}, {and} \bibinfo{person}{M Calvert}.}
  \bibinfo{year}{[n.\,d.]}\natexlab{}.
\newblock \bibinfo{title}{Key considerations to reduce or address respondent
  burden in patient-reported outcome (PRO) data collection. Nat Commun. 2022;
  13 (1): 6026}.
\newblock


\bibitem[Chen et~al\mbox{.}(2024)]%
        {chen2024unified}
\bibfield{author}{\bibinfo{person}{Boqi Chen}, \bibinfo{person}{Junier Oliva},
  {and} \bibinfo{person}{Marc Niethammer}.} \bibinfo{year}{2024}\natexlab{}.
\newblock \showarticletitle{A unified model for longitudinal multi-modal
  multi-view prediction with missingness}. In
  \bibinfo{booktitle}{\emph{International Conference on Medical Image Computing
  and Computer-Assisted Intervention}}. \bibinfo{pages}{410--420}.
\newblock


\bibitem[Cook et~al\mbox{.}(2025)]%
        {cook2025understanding}
\bibfield{author}{\bibinfo{person}{Diane Cook}, \bibinfo{person}{Aiden Walker},
  \bibinfo{person}{Bryan Minor}, \bibinfo{person}{Catherine Luna},
  \bibinfo{person}{Sarah~Tomaszewski Farias}, \bibinfo{person}{Lisa Wiese},
  \bibinfo{person}{Raven Weaver}, \bibinfo{person}{Maureen
  Schmitter-Edgecombe}, {et~al\mbox{.}}} \bibinfo{year}{2025}\natexlab{}.
\newblock \showarticletitle{Understanding the Relationship Between Ecological
  Momentary Assessment Methods, Sensed Behavior, and Responsiveness:
  Cross-Study Analysis}.
\newblock \bibinfo{journal}{\emph{JMIR mHealth and uHealth}}
  \bibinfo{volume}{13}, \bibinfo{number}{1} (\bibinfo{year}{2025}),
  \bibinfo{pages}{e57018}.
\newblock


\bibitem[Covert et~al\mbox{.}(2023)]%
        {covert2023learning}
\bibfield{author}{\bibinfo{person}{Ian~Connick Covert}, \bibinfo{person}{Wei
  Qiu}, \bibinfo{person}{Mingyu Lu}, \bibinfo{person}{Na~Yoon Kim},
  \bibinfo{person}{Nathan~J White}, {and} \bibinfo{person}{Su-In Lee}.}
  \bibinfo{year}{2023}\natexlab{}.
\newblock \showarticletitle{Learning to maximize mutual information for dynamic
  feature selection}. In \bibinfo{booktitle}{\emph{International Conference on
  Machine Learning}}. PMLR, \bibinfo{pages}{6424--6447}.
\newblock


\bibitem[Gadgil et~al\mbox{.}(2024)]%
        {gadgil2023estimating}
\bibfield{author}{\bibinfo{person}{Soham Gadgil}, \bibinfo{person}{Ian~Connick
  Covert}, {and} \bibinfo{person}{Su-In Lee}.} \bibinfo{year}{2024}\natexlab{}.
\newblock \showarticletitle{Estimating Conditional Mutual Information for
  Dynamic Feature Selection}. In \bibinfo{booktitle}{\emph{The Twelfth
  International Conference on Learning Representations}}.
\newblock


\bibitem[Ghosh and Lan(2023)]%
        {ghosh2023difa}
\bibfield{author}{\bibinfo{person}{Aritra Ghosh} {and} \bibinfo{person}{Andrew
  Lan}.} \bibinfo{year}{2023}\natexlab{}.
\newblock \showarticletitle{Difa: Differentiable feature acquisition}. In
  \bibinfo{booktitle}{\emph{Proceedings of the AAAI Conference on Artificial
  Intelligence}}, Vol.~\bibinfo{volume}{37}. \bibinfo{pages}{7705--7713}.
\newblock


\bibitem[Gong et~al\mbox{.}(2019)]%
        {gong2019icebreaker}
\bibfield{author}{\bibinfo{person}{Wenbo Gong}, \bibinfo{person}{Sebastian
  Tschiatschek}, \bibinfo{person}{Sebastian Nowozin},
  \bibinfo{person}{Richard~E Turner}, \bibinfo{person}{Jos{\'e}~Miguel
  Hern{\'a}ndez-Lobato}, {and} \bibinfo{person}{Cheng Zhang}.}
  \bibinfo{year}{2019}\natexlab{}.
\newblock \showarticletitle{Icebreaker: Element-wise efficient information
  acquisition with a bayesian deep latent gaussian model}.
\newblock \bibinfo{journal}{\emph{Advances in neural information processing
  systems}}  \bibinfo{volume}{32} (\bibinfo{year}{2019}).
\newblock


\bibitem[Houlsby et~al\mbox{.}(2011)]%
        {houlsby2011bayesian}
\bibfield{author}{\bibinfo{person}{Neil Houlsby}, \bibinfo{person}{Ferenc
  Husz{\'a}r}, \bibinfo{person}{Zoubin Ghahramani}, {and}
  \bibinfo{person}{M{\'a}t{\'e} Lengyel}.} \bibinfo{year}{2011}\natexlab{}.
\newblock \showarticletitle{Bayesian active learning for classification and
  preference learning}.
\newblock \bibinfo{journal}{\emph{arXiv preprint arXiv:1112.5745}}
  (\bibinfo{year}{2011}).
\newblock


\bibitem[Jang et~al\mbox{.}(2016)]%
        {Jang2016CategoricalRW}
\bibfield{author}{\bibinfo{person}{Eric Jang}, \bibinfo{person}{Shixiang~Shane
  Gu}, {and} \bibinfo{person}{Ben Poole}.} \bibinfo{year}{2016}\natexlab{}.
\newblock \showarticletitle{Categorical Reparameterization with
  Gumbel-Softmax}.
\newblock \bibinfo{journal}{\emph{ArXiv}}  \bibinfo{volume}{abs/1611.01144}
  (\bibinfo{year}{2016}).
\newblock
\urldef\tempurl%
\url{https://api.semanticscholar.org/CorpusID:2428314}
\showURL{%
\tempurl}


\bibitem[Janisch et~al\mbox{.}(2020)]%
        {janisch2020classification}
\bibfield{author}{\bibinfo{person}{Jarom{\'\i}r Janisch},
  \bibinfo{person}{Tom{\'a}{\v{s}} Pevn{\`y}}, {and} \bibinfo{person}{Viliam
  Lis{\`y}}.} \bibinfo{year}{2020}\natexlab{}.
\newblock \showarticletitle{Classification with costly features as a sequential
  decision-making problem}.
\newblock \bibinfo{journal}{\emph{Machine Learning}} \bibinfo{volume}{109},
  \bibinfo{number}{8} (\bibinfo{year}{2020}), \bibinfo{pages}{1587--1615}.
\newblock


\bibitem[Kellgren et~al\mbox{.}(1957)]%
        {kellgren1957radiological}
\bibfield{author}{\bibinfo{person}{Jonas~H Kellgren}, \bibinfo{person}{JS
  Lawrence}, {et~al\mbox{.}}} \bibinfo{year}{1957}\natexlab{}.
\newblock \showarticletitle{Radiological assessment of osteo-arthrosis}.
\newblock \bibinfo{journal}{\emph{Ann Rheum Dis}} \bibinfo{volume}{16},
  \bibinfo{number}{4} (\bibinfo{year}{1957}), \bibinfo{pages}{494--502}.
\newblock


\bibitem[Khaire and Dhanalakshmi(2022)]%
        {khaire2022stability}
\bibfield{author}{\bibinfo{person}{Utkarsh~Mahadeo Khaire} {and}
  \bibinfo{person}{R Dhanalakshmi}.} \bibinfo{year}{2022}\natexlab{}.
\newblock \showarticletitle{Stability of feature selection algorithm: A
  review}.
\newblock \bibinfo{journal}{\emph{Journal of King Saud University-Computer and
  Information Sciences}} \bibinfo{volume}{34}, \bibinfo{number}{4}
  (\bibinfo{year}{2022}), \bibinfo{pages}{1060--1073}.
\newblock


\bibitem[Kidger et~al\mbox{.}(2020)]%
        {kidger2020neural}
\bibfield{author}{\bibinfo{person}{Patrick Kidger}, \bibinfo{person}{James
  Morrill}, \bibinfo{person}{James Foster}, {and} \bibinfo{person}{Terry
  Lyons}.} \bibinfo{year}{2020}\natexlab{}.
\newblock \showarticletitle{Neural controlled differential equations for
  irregular time series}.
\newblock \bibinfo{journal}{\emph{Advances in Neural Information Processing
  Systems}}  \bibinfo{volume}{33} (\bibinfo{year}{2020}),
  \bibinfo{pages}{6696--6707}.
\newblock


\bibitem[Konyushkova et~al\mbox{.}(2017)]%
        {konyushkova2017learning}
\bibfield{author}{\bibinfo{person}{Ksenia Konyushkova},
  \bibinfo{person}{Raphael Sznitman}, {and} \bibinfo{person}{Pascal Fua}.}
  \bibinfo{year}{2017}\natexlab{}.
\newblock \showarticletitle{Learning active learning from data}.
\newblock \bibinfo{journal}{\emph{Advances in neural information processing
  systems}}  \bibinfo{volume}{30} (\bibinfo{year}{2017}).
\newblock


\bibitem[Kossen et~al\mbox{.}(2023)]%
        {kossen2022active}
\bibfield{author}{\bibinfo{person}{Jannik Kossen},
  \bibinfo{person}{C{\u{a}}t{\u{a}}lina Cangea}, \bibinfo{person}{Eszter
  V{\'e}rtes}, \bibinfo{person}{Andrew Jaegle}, \bibinfo{person}{Viorica
  Patraucean}, \bibinfo{person}{Ira Ktena}, \bibinfo{person}{Nenad Tomasev},
  {and} \bibinfo{person}{Danielle Belgrave}.} \bibinfo{year}{2023}\natexlab{}.
\newblock \showarticletitle{Active Acquisition for Multimodal Temporal Data: A
  Challenging Decision-Making Task}.
\newblock \bibinfo{journal}{\emph{Transactions on Machine Learning Research}}
  (\bibinfo{year}{2023}).
\newblock


\bibitem[Li and Oliva(2021)]%
        {li2021active}
\bibfield{author}{\bibinfo{person}{Yang Li} {and} \bibinfo{person}{Junier
  Oliva}.} \bibinfo{year}{2021}\natexlab{}.
\newblock \showarticletitle{Active feature acquisition with generative
  surrogate models}. In \bibinfo{booktitle}{\emph{International conference on
  machine learning}}. PMLR, \bibinfo{pages}{6450--6459}.
\newblock


\bibitem[Ma et~al\mbox{.}(2019)]%
        {ma2018eddi}
\bibfield{author}{\bibinfo{person}{Chao Ma}, \bibinfo{person}{Sebastian
  Tschiatschek}, \bibinfo{person}{Konstantina Palla},
  \bibinfo{person}{Jose~Miguel Hernandez-Lobato}, \bibinfo{person}{Sebastian
  Nowozin}, {and} \bibinfo{person}{Cheng Zhang}.}
  \bibinfo{year}{2019}\natexlab{}.
\newblock \showarticletitle{EDDI: Efficient Dynamic Discovery of High-Value
  Information with Partial VAE}. In \bibinfo{booktitle}{\emph{International
  Conference on Machine Learning}}. PMLR, \bibinfo{pages}{4234--4243}.
\newblock


\bibitem[MacKay(1992)]%
        {mackay1992information}
\bibfield{author}{\bibinfo{person}{David~JC MacKay}.}
  \bibinfo{year}{1992}\natexlab{}.
\newblock \showarticletitle{Information-based objective functions for active
  data selection}.
\newblock \bibinfo{journal}{\emph{Neural computation}} \bibinfo{volume}{4},
  \bibinfo{number}{4} (\bibinfo{year}{1992}), \bibinfo{pages}{590--604}.
\newblock


\bibitem[Maddison et~al\mbox{.}(2016)]%
        {Maddison2016TheCD}
\bibfield{author}{\bibinfo{person}{Chris~J. Maddison}, \bibinfo{person}{Andriy
  Mnih}, {and} \bibinfo{person}{Yee~Whye Teh}.}
  \bibinfo{year}{2016}\natexlab{}.
\newblock \showarticletitle{The Concrete Distribution: A Continuous Relaxation
  of Discrete Random Variables}.
\newblock \bibinfo{journal}{\emph{ArXiv}}  \bibinfo{volume}{abs/1611.00712}
  (\bibinfo{year}{2016}).
\newblock
\urldef\tempurl%
\url{https://api.semanticscholar.org/CorpusID:14307651}
\showURL{%
\tempurl}


\bibitem[McConnell et~al\mbox{.}(2001)]%
        {mcconnell2001western}
\bibfield{author}{\bibinfo{person}{Sara McConnell}, \bibinfo{person}{Pamela
  Kolopack}, {and} \bibinfo{person}{Aileen~M Davis}.}
  \bibinfo{year}{2001}\natexlab{}.
\newblock \showarticletitle{The Western Ontario and McMaster Universities
  Osteoarthritis Index (WOMAC): a review of its utility and measurement
  properties}.
\newblock \bibinfo{journal}{\emph{Arthritis Care \& Research: Official Journal
  of the American College of Rheumatology}} \bibinfo{volume}{45},
  \bibinfo{number}{5} (\bibinfo{year}{2001}), \bibinfo{pages}{453--461}.
\newblock


\bibitem[Miao and Niu(2016)]%
        {miao2016survey}
\bibfield{author}{\bibinfo{person}{Jianyu Miao} {and} \bibinfo{person}{Lingfeng
  Niu}.} \bibinfo{year}{2016}\natexlab{}.
\newblock \showarticletitle{A survey on feature selection}.
\newblock \bibinfo{journal}{\emph{Procedia computer science}}
  \bibinfo{volume}{91} (\bibinfo{year}{2016}), \bibinfo{pages}{919--926}.
\newblock


\bibitem[Nguyen et~al\mbox{.}(2024)]%
        {nguyen2024active}
\bibfield{author}{\bibinfo{person}{Khanh Nguyen}, \bibinfo{person}{Huy~Hoang
  Nguyen}, \bibinfo{person}{Egor Panfilov}, {and} \bibinfo{person}{Aleksei
  Tiulpin}.} \bibinfo{year}{2024}\natexlab{}.
\newblock \showarticletitle{Active Sensing of Knee Osteoarthritis Progression
  with Reinforcement Learning}.
\newblock \bibinfo{journal}{\emph{arXiv preprint arXiv:2408.02349}}
  (\bibinfo{year}{2024}).
\newblock


\bibitem[Norcliffe et~al\mbox{.}(2025)]%
        {norcliffe2025stochastic}
\bibfield{author}{\bibinfo{person}{Alexander Norcliffe},
  \bibinfo{person}{Changhee Lee}, \bibinfo{person}{Fergus Imrie},
  \bibinfo{person}{Mihaela Van Der~Schaar}, {and} \bibinfo{person}{Pietro
  Li{\`o}}.} \bibinfo{year}{2025}\natexlab{}.
\newblock \showarticletitle{Stochastic Encodings for Active Feature
  Acquisition}.
\newblock \bibinfo{journal}{\emph{arXiv preprint arXiv:2508.01957}}
  (\bibinfo{year}{2025}).
\newblock


\bibitem[O’Bryant et~al\mbox{.}(2010)]%
        {o2010validation}
\bibfield{author}{\bibinfo{person}{Sid~E O’Bryant}, \bibinfo{person}{Laura~H
  Lacritz}, \bibinfo{person}{James Hall}, \bibinfo{person}{Stephen~C Waring},
  \bibinfo{person}{Wenyaw Chan}, \bibinfo{person}{Zeina~G Khodr},
  \bibinfo{person}{Paul~J Massman}, \bibinfo{person}{Valerie Hobson}, {and}
  \bibinfo{person}{C~Munro Cullum}.} \bibinfo{year}{2010}\natexlab{}.
\newblock \showarticletitle{Validation of the new interpretive guidelines for
  the clinical dementia rating scale sum of boxes score in the national
  Alzheimer's coordinating center database}.
\newblock \bibinfo{journal}{\emph{Archives of neurology}} \bibinfo{volume}{67},
  \bibinfo{number}{6} (\bibinfo{year}{2010}), \bibinfo{pages}{746--749}.
\newblock


\bibitem[O’Bryant et~al\mbox{.}(2008)]%
        {o2008staging}
\bibfield{author}{\bibinfo{person}{Sid~E O’Bryant},
  \bibinfo{person}{Stephen~C Waring}, \bibinfo{person}{C~Munro Cullum},
  \bibinfo{person}{James Hall}, \bibinfo{person}{Laura Lacritz},
  \bibinfo{person}{Paul~J Massman}, \bibinfo{person}{Philip~J Lupo},
  \bibinfo{person}{Joan~S Reisch}, \bibinfo{person}{Rachelle Doody},
  \bibinfo{person}{Texas Alzheimer's~Research Consortium}, {et~al\mbox{.}}}
  \bibinfo{year}{2008}\natexlab{}.
\newblock \showarticletitle{Staging dementia using Clinical Dementia Rating
  Scale Sum of Boxes scores: a Texas Alzheimer's research consortium study}.
\newblock \bibinfo{journal}{\emph{Archives of neurology}} \bibinfo{volume}{65},
  \bibinfo{number}{8} (\bibinfo{year}{2008}), \bibinfo{pages}{1091--1095}.
\newblock


\bibitem[Petersen et~al\mbox{.}(2010)]%
        {petersen2010alzheimer}
\bibfield{author}{\bibinfo{person}{Ronald~Carl Petersen},
  \bibinfo{person}{Paul~S Aisen}, \bibinfo{person}{Laurel~A Beckett},
  \bibinfo{person}{Michael~C Donohue}, \bibinfo{person}{Anthony~Collins Gamst},
  \bibinfo{person}{Danielle~J Harvey}, \bibinfo{person}{CR Jack~Jr},
  \bibinfo{person}{William~J Jagust}, \bibinfo{person}{Leslie~M Shaw},
  \bibinfo{person}{Arthur~W Toga}, {et~al\mbox{.}}}
  \bibinfo{year}{2010}\natexlab{}.
\newblock \showarticletitle{Alzheimer's disease Neuroimaging Initiative (ADNI)
  clinical characterization}.
\newblock \bibinfo{journal}{\emph{Neurology}} \bibinfo{volume}{74},
  \bibinfo{number}{3} (\bibinfo{year}{2010}), \bibinfo{pages}{201--209}.
\newblock


\bibitem[Qin et~al\mbox{.}(2024)]%
        {qin2024risk}
\bibfield{author}{\bibinfo{person}{Yuchao Qin}, \bibinfo{person}{Mihaela
  van~der Schaar}, {and} \bibinfo{person}{Changhee Lee}.}
  \bibinfo{year}{2024}\natexlab{}.
\newblock \showarticletitle{Risk-averse active sensing for timely outcome
  prediction under cost pressure}.
\newblock \bibinfo{journal}{\emph{Advances in Neural Information Processing
  Systems}}  \bibinfo{volume}{36} (\bibinfo{year}{2024}).
\newblock


\bibitem[Ringwald et~al\mbox{.}(2025a)]%
        {Ringwald2025CommonAU}
\bibfield{author}{\bibinfo{person}{Whitney~R. Ringwald},
  \bibinfo{person}{Kasey~G. Creswell}, \bibinfo{person}{Carissa~A. Low},
  \bibinfo{person}{Afsaneh Doryab}, \bibinfo{person}{Tammy Chung},
  \bibinfo{person}{Junier Oliva}, \bibinfo{person}{Zachary~F Fisher},
  \bibinfo{person}{Kathleen~M Gates}, {and} \bibinfo{person}{Aidan G.~C.
  Wright}.} \bibinfo{year}{2025}\natexlab{a}.
\newblock \showarticletitle{Common and uncommon risky drinking patterns in
  young adulthood uncovered by person-specific computational modeling.}
\newblock \bibinfo{journal}{\emph{Psychology of addictive behaviors : journal
  of the Society of Psychologists in Addictive Behaviors}}
  (\bibinfo{year}{2025}).
\newblock
\urldef\tempurl%
\url{https://api.semanticscholar.org/CorpusID:275442415}
\showURL{%
\tempurl}


\bibitem[Ringwald et~al\mbox{.}(2025b)]%
        {Ringwald2025DoYF}
\bibfield{author}{\bibinfo{person}{Whitney~R. Ringwald},
  \bibinfo{person}{Colin~E. Vize}, {and} \bibinfo{person}{Aidan G.~C. Wright}.}
  \bibinfo{year}{2025}\natexlab{b}.
\newblock \showarticletitle{Do you feel what I feel? The relation between
  congruence of perceived affect and self-reported empathy in daily life social
  situations.}
\newblock \bibinfo{journal}{\emph{Emotion}} (\bibinfo{year}{2025}).
\newblock
\urldef\tempurl%
\url{https://api.semanticscholar.org/CorpusID:278234840}
\showURL{%
\tempurl}


\bibitem[Rolstad et~al\mbox{.}(2011)]%
        {rolstad2011response}
\bibfield{author}{\bibinfo{person}{Sindre Rolstad}, \bibinfo{person}{John
  Adler}, {and} \bibinfo{person}{Anna Ryd{\'e}n}.}
  \bibinfo{year}{2011}\natexlab{}.
\newblock \showarticletitle{Response burden and questionnaire length: is
  shorter better? A review and meta-analysis}.
\newblock \bibinfo{journal}{\emph{Value in Health}} \bibinfo{volume}{14},
  \bibinfo{number}{8} (\bibinfo{year}{2011}), \bibinfo{pages}{1101--1108}.
\newblock


\bibitem[Ross et~al\mbox{.}(2011)]%
        {ross2011reduction}
\bibfield{author}{\bibinfo{person}{St{\'e}phane Ross},
  \bibinfo{person}{Geoffrey Gordon}, {and} \bibinfo{person}{Drew Bagnell}.}
  \bibinfo{year}{2011}\natexlab{}.
\newblock \showarticletitle{A reduction of imitation learning and structured
  prediction to no-regret online learning}. In
  \bibinfo{booktitle}{\emph{Proceedings of the fourteenth international
  conference on artificial intelligence and statistics}}. JMLR Workshop and
  Conference Proceedings, \bibinfo{pages}{627--635}.
\newblock


\bibitem[Saar-Tsechansky et~al\mbox{.}(2009)]%
        {saar2009active}
\bibfield{author}{\bibinfo{person}{Maytal Saar-Tsechansky},
  \bibinfo{person}{Prem Melville}, {and} \bibinfo{person}{Foster Provost}.}
  \bibinfo{year}{2009}\natexlab{}.
\newblock \showarticletitle{Active feature-value acquisition}.
\newblock \bibinfo{journal}{\emph{Management Science}} \bibinfo{volume}{55},
  \bibinfo{number}{4} (\bibinfo{year}{2009}), \bibinfo{pages}{664--684}.
\newblock


\bibitem[Sheng and Ling(2006)]%
        {sheng2006feature}
\bibfield{author}{\bibinfo{person}{Victor~S Sheng} {and}
  \bibinfo{person}{Charles~X Ling}.} \bibinfo{year}{2006}\natexlab{}.
\newblock \showarticletitle{Feature value acquisition in testing: a sequential
  batch test algorithm}. In \bibinfo{booktitle}{\emph{Proceedings of the 23rd
  international conference on Machine learning}}. \bibinfo{pages}{809--816}.
\newblock


\bibitem[Shiffman et~al\mbox{.}(2008)]%
        {shiffman2008ecological}
\bibfield{author}{\bibinfo{person}{Saul Shiffman}, \bibinfo{person}{Arthur~A
  Stone}, {and} \bibinfo{person}{Michael~R Hufford}.}
  \bibinfo{year}{2008}\natexlab{}.
\newblock \showarticletitle{Ecological momentary assessment}.
\newblock \bibinfo{journal}{\emph{Annu. Rev. Clin. Psychol.}}
  \bibinfo{volume}{4}, \bibinfo{number}{1} (\bibinfo{year}{2008}),
  \bibinfo{pages}{1--32}.
\newblock


\bibitem[Shim et~al\mbox{.}(2018)]%
        {shim2018joint}
\bibfield{author}{\bibinfo{person}{Hajin Shim}, \bibinfo{person}{Sung~Ju
  Hwang}, {and} \bibinfo{person}{Eunho Yang}.} \bibinfo{year}{2018}\natexlab{}.
\newblock \showarticletitle{Joint active feature acquisition and classification
  with variable-size set encoding}.
\newblock \bibinfo{journal}{\emph{Advances in neural information processing
  systems}}  \bibinfo{volume}{31} (\bibinfo{year}{2018}).
\newblock


\bibitem[Swinckels et~al\mbox{.}(2024)]%
        {swinckels2024use}
\bibfield{author}{\bibinfo{person}{Laura Swinckels}, \bibinfo{person}{Frank~C
  Bennis}, \bibinfo{person}{Kirsten~A Ziesemer}, \bibinfo{person}{Janneke~FM
  Scheerman}, \bibinfo{person}{Harmen Bijwaard}, \bibinfo{person}{Ander de
  Keijzer}, {and} \bibinfo{person}{Josef~Jan Bruers}.}
  \bibinfo{year}{2024}\natexlab{}.
\newblock \showarticletitle{The use of deep learning and machine learning on
  longitudinal electronic health records for the early detection and prevention
  of diseases: scoping review}.
\newblock \bibinfo{journal}{\emph{Journal of medical Internet research}}
  \bibinfo{volume}{26} (\bibinfo{year}{2024}), \bibinfo{pages}{e48320}.
\newblock


\bibitem[Valancius et~al\mbox{.}(2024)]%
        {valancius2024acquisition}
\bibfield{author}{\bibinfo{person}{Michael Valancius}, \bibinfo{person}{Maxwell
  Lennon}, {and} \bibinfo{person}{Junier Oliva}.}
  \bibinfo{year}{2024}\natexlab{}.
\newblock \showarticletitle{Acquisition Conditioned Oracle for Nongreedy Active
  Feature Acquisition}. In \bibinfo{booktitle}{\emph{International Conference
  on Machine Learning}}. PMLR, \bibinfo{pages}{48957--48975}.
\newblock


\bibitem[Venkatesh and Anuradha(2019)]%
        {venkatesh2019review}
\bibfield{author}{\bibinfo{person}{B Venkatesh} {and} \bibinfo{person}{J
  Anuradha}.} \bibinfo{year}{2019}\natexlab{}.
\newblock \showarticletitle{A review of feature selection and its methods}.
\newblock \bibinfo{journal}{\emph{Cybern. Inf. Technol}} \bibinfo{volume}{19},
  \bibinfo{number}{1} (\bibinfo{year}{2019}), \bibinfo{pages}{3--26}.
\newblock


\bibitem[Yin et~al\mbox{.}(2020)]%
        {yin2020reinforcement}
\bibfield{author}{\bibinfo{person}{Haiyan Yin}, \bibinfo{person}{Yingzhen Li},
  \bibinfo{person}{Sinno~Jialin Pan}, \bibinfo{person}{Cheng Zhang}, {and}
  \bibinfo{person}{Sebastian Tschiatschek}.} \bibinfo{year}{2020}\natexlab{}.
\newblock \showarticletitle{Reinforcement learning with efficient active
  feature acquisition}.
\newblock \bibinfo{journal}{\emph{arXiv preprint arXiv:2011.00825}}
  (\bibinfo{year}{2020}).
\newblock


\bibitem[Yoon et~al\mbox{.}(2019)]%
        {yoon2019asac}
\bibfield{author}{\bibinfo{person}{Jinsung Yoon}, \bibinfo{person}{James
  Jordon}, {and} \bibinfo{person}{Mihaela Schaar}.}
  \bibinfo{year}{2019}\natexlab{}.
\newblock \showarticletitle{ASAC: Active sensing using Actor-Critic models}. In
  \bibinfo{booktitle}{\emph{Machine Learning for Healthcare Conference}}. PMLR,
  \bibinfo{pages}{451--473}.
\newblock


\end{thebibliography}

\clearpage
\appendix
\section*{Appendix}

\section{Additional Results for Experiments and Ablations}
\label{appendix:add_results}
\subsection{AUPRC Results}
\begin{figure*}[h]
\centering
\includegraphics[width=1\linewidth]{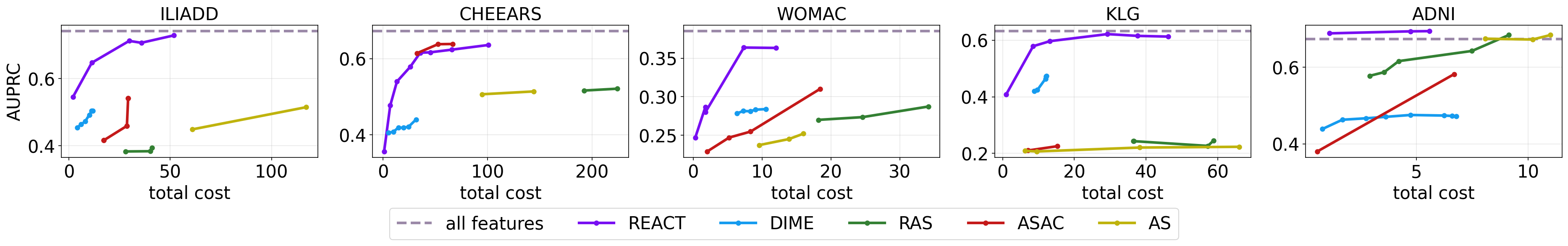}
\caption{AUPRC/total cost of models 
  across various average acquisition costs (budgets) on test data. The dashed line is evaluating the pretrained classifier from REACT with all features available.}
\label{fig:performance_cost_AUPRC}

\end{figure*}

\begin{figure*}[t]
    \centering
    \includegraphics[width=1\linewidth]{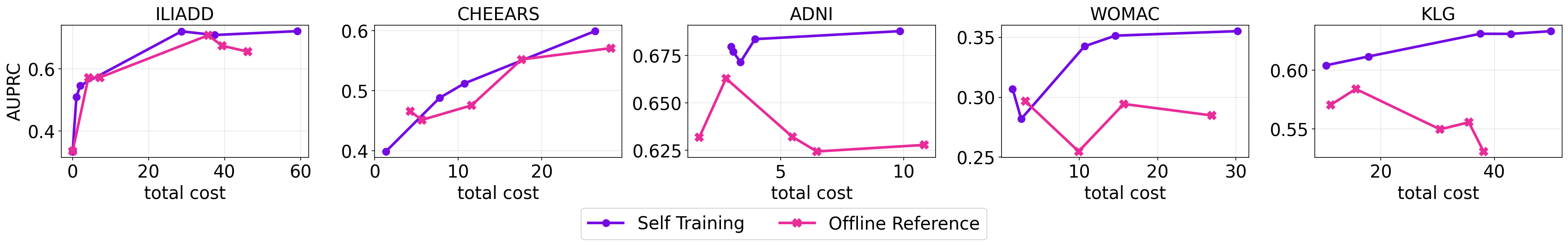}
    \caption{Ablation study of the self-iterative training for \OurMethod (Alg. \ref{alg:training}). We compare AUPRC/total cost when planner $\pi_\theta$ trained with (i) self-iterative training from random initialization (Self Training) and (ii) trained using only offline reference states (Offline Reference).}
    \label{fig:ABL_warmup_auprc}
\end{figure*}

\begin{figure*}[t]
    \centering
    \includegraphics[width=1\linewidth]{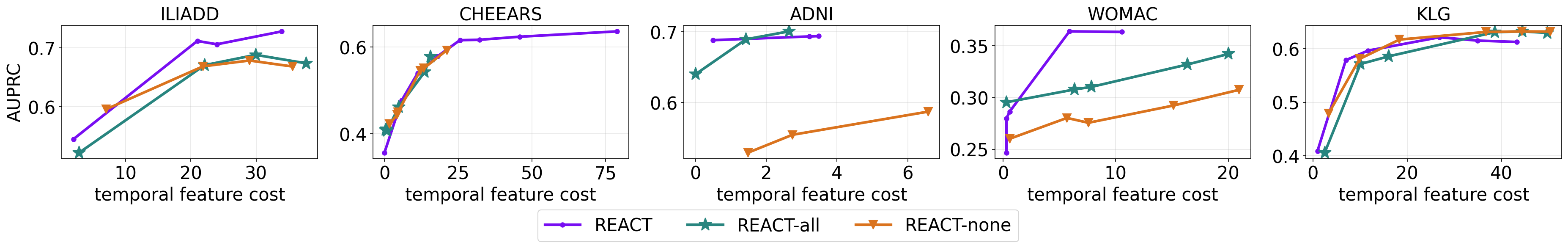}
    \caption{Ablation study on how the learned context selector affects the temporal feature acquisition. AUPRC of \OurMethod with learned context descriptors compared to when context descriptors are all acquired (\OurMethod-all) or not at all acquired (\OurMethod-none) across datasets.}
    \label{fig:ABL_contex_auprc}
\end{figure*}

We provide the additional AUPRC results for the cost/performance trade-off experiment presented in the main text. Fig.~\ref{fig:performance_cost_AUPRC} is the experiment results, Fig.~\ref{fig:ABL_contex_auprc} is the context selector ablation results, and Fig.~\ref{fig:ABL_warmup_auprc} is the oracle vs self-training ablation results.

\subsection{Warmup Ablation}
\label{appendix:warmup}

\begin{figure}[H]
    \centering
    \includegraphics[width=0.7\linewidth]{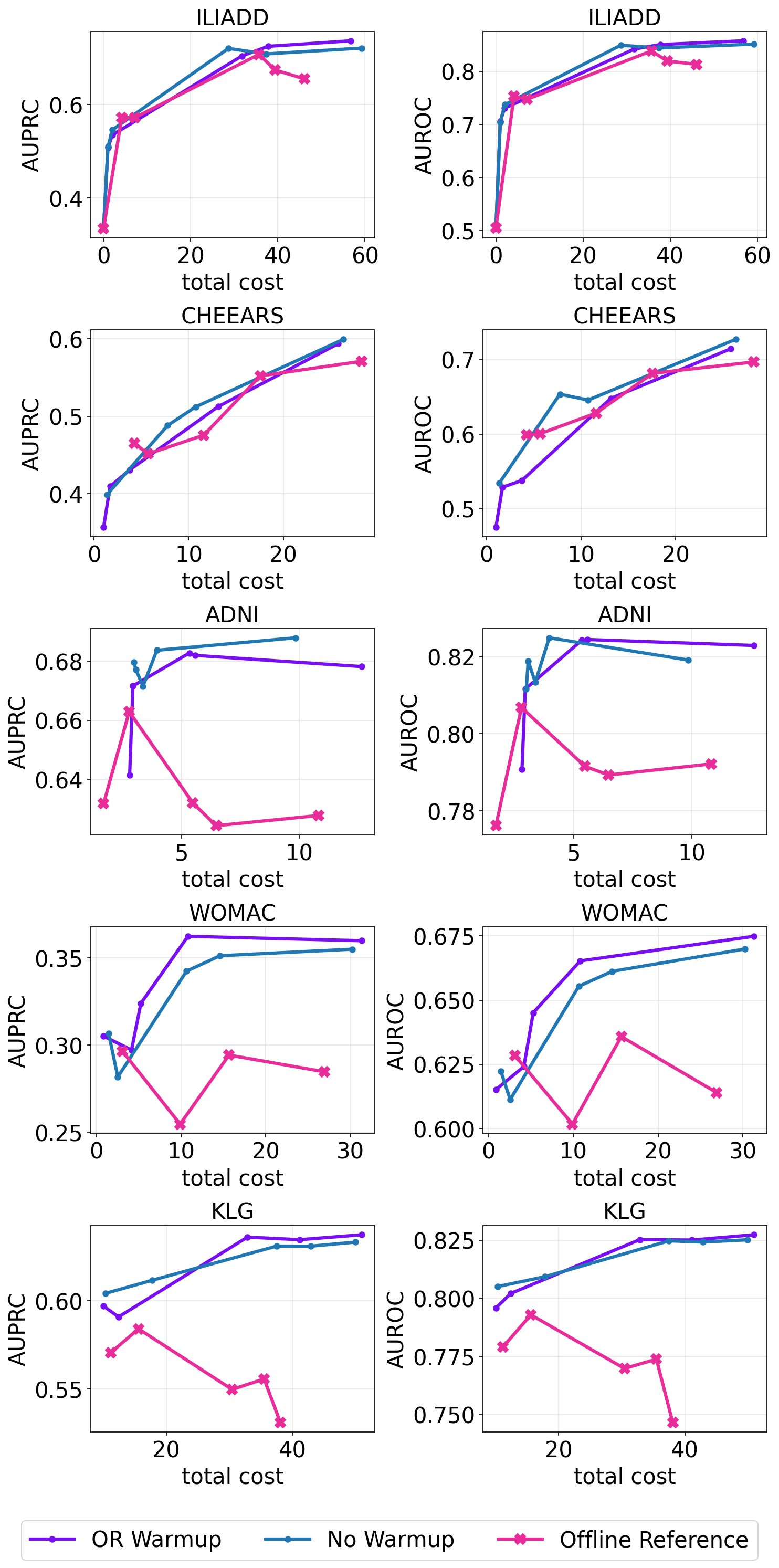}
    \caption{Ablation on States Seen During Policy Training.}
    \label{fig:ABL_OR_warmup}
\end{figure}

While offline references can provide states to train the planner network, they have a fundamental limitation: one trajectory per instance. Online refinement solves this via on-policy rollouts: executing $\pi_{\theta}$ generates diverse acquisition, exposing the policy to more states. We ablate on the type of states seen during training to evaluate if training using states output by the planner network $\pi_{\theta}$ helps generalization to test data. We ablate on three strategies: (1) Offline Reference (OR) Warmup (\OurMethod), (2) No Warmup (pure online), and (3) Offline Reference Only. We will detail later in this section how to create the reference plan datasets for warmup. Fig.~\ref{fig:ABL_OR_warmup} shows OR Warmup achieves comparable cost-performance tradeoffs with No Warmup on most datasets and better on WOMAC. 

\paragraph{\textbf{Offline Reference Warmup}}
In this section, we detail how to create the dataset of reference plans, which is then used to warmup the planner $\pi_\theta$. On the high-level, because training trajectories are fully observed, we can evaluate candidate acquisition plans under the same accuracy--cost tradeoff used in our objective and select the best-scoring plan for each training instance. We then initialize training episodes with these instance-specific reference plans, rather than random states, to provide a stronger starting point for the planner before subsequent self-iterative refinement with on-policy rollouts.

Let $m_s \in \{0,1\}^{d_s}$ denote the contextual acquisition mask, and let $M \in \{0,1\}^{T \times d}$ denote the temporal acquisition plan, with row $m_t \in \{0,1\}^d$ corresponding to the acquired temporal features at time $t$. Given a training trajectory $(s,x,y) \in \mathcal{D}_{\text{train}}$, where $y = (y_1,\dots,y_T)$, we define $\tilde{s} = m_s \odot s$ and $\tilde{x}_t = m_t \odot x_t$ for $t=1,\dots,T$. Let
\begin{equation}
H_t(M) = (\tilde{x}_1,\dots,\tilde{x}_t,0,\dots,0)
\end{equation}
denote the masked temporal history available up to time $t$. We then score a candidate plan $(m_s, M)$ using the plug-in objective
\begin{equation}
\label{eq:plugin_objective}
\scalebox{0.9}{%
$J_{\lambda}(m_s, M; x,s,y)
=
\sum_{t=1}^{T}
L_{\mathrm{pred}}\!\left(f_{\phi}(H_t(M), \tilde{s}, t),\, y_t\right)
+
\lambda \left(
c_s^{\top} m_s
+
\sum_{t=1}^{T} c_x^{\top} m_t
\right)$%
}
\end{equation}
Because $(s,x,y)$ is fully observed during training, $J_{\lambda}$ is directly computable and can therefore be used to compare candidate discrete acquisition plans, as described below.

For each training instance $i$, we generate a set of candidate acquisition plans ($K=1000$ in our experiments)
\begin{equation}
\mathrm{Candidate}^i = \{(m_s^{(k)}, M^{(k)})\}_{k=1}^{K},
\end{equation}
and select the best plan, referred to as the reference plan, by
\begin{equation}
\label{eq:oracle_plan}
(m_s^{\ast}, M^{\ast})
\in
\argmin_{(m_s,M)\in \mathrm{Candidate}^i}
J_{\lambda}(m_s, M; x^i, s^i, y^i).
\end{equation}
In practice, rather than performing an exhaustive search over $\mathrm{Candidate}^i$, we construct the candidate set by sampling discrete plans, e.g., uniformly at random.  After obtaining reference plans for all training instances, we form a reference dataset
\begin{equation}
\mathcal{D}_{\mathrm{ref}}
=
\left\{
\bigl((s^i, x^i, y^i), (m_s^{\ast}, M^{\ast})\bigr)
\right\}_{i=1}^{N},
\end{equation}
which maps each fully observed training trajectory to its best-scoring acquisition plan and is used to warm-start the planner $\pi$. We use $\mathcal{D}_{\mathrm{ref}}$ to warm-start the policy parameters $\alpha$ and $\theta$ of the planner $\pi_\theta$ by minimizing the prediction loss from the classifier when the planner sees the states from reference masks.

\section{Baseline Implementations}
\label{appendix:baseline}
We evaluate our proposed framework against several strong baseline methods for traditional AFA and LAFA settings. Importantly, existing baselines do not natively distinguish the onboarding context $s$ from the temporal measurements $x_t$. To ensure a fair comparison, we restructure the observation space for all baseline models by concatenating both feature types at every step, obtaining the new state vector $x'_t = [s, x_t] \in \mathbb{R}^{d_s+d}$. Thus, this grants the baselines flexibility where they can acquire missed contextual descriptors earlier. Consequently, the baseline policies have to discover that context should be acquired early, while actively learning to avoid remeasuring (and cost penalty $c_s$) the same context data in subsequent visits. 


To ensure a fair comparison, we evaluated all baselines by adapting the authors' official implementations. For ASAC, RAS, and AS, we use the implementation available at \url{https://github.com/yvchao/cvar_sensing}, which is under the BSD-3-Clause license. For DIME, built on top of the official implementation available at \url{https://github.com/suinleelab/DIME}, we adapt the method to the longitudinal setting by restricting the acquisition to current or future timesteps.


Following \cite{qin2024risk}, ASAC, RAS, and AS share the same neural CDE predictor \cite{kidger2020neural}. Following the released code, the drop rate $p \in \{0.0, 0.3, 0.5, 0.7\}$ of the auxiliary observation strategy $\pi_0$ is treated as a hyperparameter and selected according to the outcome predictor's average accuracy over multiple randomly masked evaluation sets generated by $\pi_0$. The selected drop rates are:
\[
p =
\begin{cases}
\texttt{0.7} & \text{ILIADD},\\
\texttt{0.7} & \text{CHEEARS},\\
\texttt{0.3} & \text{ADNI},\\
\texttt{0.3} & \text{WOMAC},\\
\texttt{0.0} & \text{KLG}.
\end{cases}
\]

\paragraph{\textbf{ASAC \cite{yoon2019asac}}}
For ASAC, we tune the acquisition-cost coefficient $\mu$ on the validation set. The search grids are:
\[
\mu \in
\begin{cases}
\{\texttt{0.02, 0.005, 0.002}\} & \text{ILIADD},\\
\{\texttt{0.01, 0.005, 0.003, 0.002}\} & \text{CHEEARS},\\
\{\texttt{0.02, 0.01, 0.001}\} & \text{ADNI},\\
\{\texttt{0.1, 0.01, 0.005, 0.001}\} & \text{WOMAC},\\
\{\texttt{0.00175, 0.00125}\} & \text{KLG}.
\end{cases}
\]
Note that sweeping the $\mu$ parameter yields a diverse set of ASAC policies, each corresponding to a different average acquisition cost.

\paragraph{\textbf{RAS \cite{qin2024risk}}}
For RAS, we specify the acquisition interval range $(\Delta_{\min}, \Delta_{\max}) = (0.5, 1.5)$ to all datasets. We further tune the diagnostic-error coefficient $\gamma_{\text{RAS}}$ on the validation set via grid search:
\[
\gamma_{\text{RAS}} \in
\begin{cases}
\{\texttt{5, 10, 15}\} & \text{ILIADD},\\
\{\texttt{5, 10, 15}\} & \text{CHEEARS},\\
\{\texttt{50, 75, 100, 175}\} & \text{ADNI},\\
\{\texttt{500, 750, 1000, 1250}\} & \text{WOMAC},\\
\{\texttt{500, 750, 1000, 1250}\} & \text{KLG}.
\end{cases}
\]
The final selected values of $\gamma_{\text{RAS}}$ depend on the target acquisition budget. For all datasets, we use a tail-risk quantile of $\texttt{0.1}$, an invalid-visit penalty of $\texttt{10}$, and a discount factor of $\texttt{0.99}$, following the authors’ defaults.

\paragraph{\textbf{AS \cite{qin2024risk}}}
For AS, we use the same minimum and maximum acquisition-interval constraints as in the RAS setting. In addition, we tune a fixed acquisition interval $\tilde{\Delta}$ for each dataset:
\[
\tilde{\Delta} \in
\begin{cases}
\{\texttt{0.5, 1.0, 1.5}\} & \text{ILIADD},\\
\{\texttt{1.0, 1.5, 2.0}\} & \text{CHEEARS},\\
\{\texttt{0.2, 0.4, 0.5, 1.0, 1.5}\} & \text{ADNI},\\
\{\texttt{1.0, 1.5, 2.0}\} & \text{WOMAC},\\
\{\texttt{0.2, 0.4, 0.5, 1.0, 1.5}\} & \text{KLG}.
\end{cases}
\]
As with RAS, different final values of $\tilde{\Delta}$ may be reported for different acquisition budgets.

\paragraph{\textbf{DIME \cite{covert2023learning}}}
We extend DIME to the longitudinal setting while preserving its original greedy acquisition mechanism. In our implementation, DIME is restricted to selecting features from the current or future time points only. Since DIME acquires one feature at a time, it may make repeated selections within the same time step before moving forward in time. To keep the comparison controlled, both the prediction network and value network of DIME share the same architecture
as \OurMethod.

\section{Additional Implementation and Training Details on \OurMethod}
\label{appendix:experiment}
Details architecture and training procedure for \OurMethod are as follow:
\begin{itemize}
    \item predictor: 3-layer MLP with ReLU activations;  64 for all hidden layer size; pre-trained with cross entropy loss, dropout with $\rho$=0.4, random masking of input where each subset size is equally likely, and early stopping on the validation set.
    \item contextual descriptor selector: logits of context descriptor size
    \item planner: 3-layer MLP with ReLU activations; hidden layer sizes are [512, 256,128]; input time indicator embedded with sinusoidal time embedding with dimension of 64.
    \item training procedure: For results in Fig.~\ref{fig:performance_cost} and Fig.~\ref{fig:performance_cost_AUPRC}, we jointly trained the above 3 components for 1000 mini batches of size 64 on all datasets, optimizing using $\mathcal{L}_{\OurMethod}$. The first 50 are warm-up batches using offline reference, and the remaining 950 are self-training.
\end{itemize}
Hyperparameters we used for \OurMethod are as follows:
\begin{itemize}
    \item cost-benefit tradeoff hyperparameter $\lambda$: in \autoref{tab:react_costs}, we report $\lambda$ we used for each cost we have in our plots.
    \item learning rate for pretraining predictor: 0.001.
    \item learning rate for training the planner and context selector: 0.001.
    \item learning rate for training predictor jointly: 0.0001.
\end{itemize}

\begin{table*}
\centering
\small
\begin{tabular}{|l|c|r|r|r|}
\hline
\textbf{Dataset} & \textbf{$\lambda$} & \textbf{Total Cost} & \textbf{Temporal Cost} & \textbf{Context Cost} \\
\hline
ADNI & 0.05   & 1.100  & 0.500  & 0.600 \\
     & 0.01   & 4.725  & 3.225  & 1.500 \\
     & 0.0001 & 5.585  & 3.485  & 2.100 \\
\hline
CHEEARS & 0.003  & 1.000   & 0.000  & 1.000  \\
        & 0.002  & 6.851   & 5.851  & 1.000  \\
        & 0.0015 & 13.275  & 11.275 & 2.000  \\
        & 0.0008 & 26.044  & 18.044 & 8.000  \\
        & 0.0006 & 35.571  & 25.571 & 10.000 \\
        & 0.0004 & 45.213  & 32.257 & 12.956 \\
        & 0.0002 & 65.754  & 45.798 & 19.956 \\
        & 0.0001 & 100.824 & 78.868 & 21.956 \\
\hline
ILIADD & 0.005  & 1.999  & 1.999  & 0.000  \\
       & 0.003  & 11.266 & 11.266 & 0.000  \\
       & 0.001  & 29.742 & 20.995 & 8.747  \\
       & 0.0005 & 35.739 & 23.993 & 11.747 \\
       & 0.0001 & 51.696 & 33.950 & 17.747 \\
\hline
KLG & 0.05   & 0.989  & 0.989  & 0.000 \\
    & 0.01   & 8.489  & 6.997  & 1.492 \\
    & 0.005  & 13.179 & 11.687 & 1.492 \\
    & 0.001  & 29.243 & 26.869 & 2.374 \\
    & 0.0005 & 37.613 & 34.941 & 2.672 \\
    & 0.0001 & 46.243 & 43.274 & 2.969 \\
\hline
WOMAC & 0.05   & 0.297  & 0.297  & 0.000 \\
      & 0.01   & 1.767  & 0.584  & 1.183 \\
      & 0.005  & 1.775  & 0.297  & 1.478 \\
      & 0.001  & 7.368  & 5.890  & 1.478 \\
      & 0.0005 & 12.037 & 10.559 & 1.478 \\
\hline
\end{tabular}
\caption{REACT hyperparameter: acquisition costs and $\lambda$ by dataset.}
\label{tab:react_costs}
\end{table*}

\section{Training and Evaluation Wall-Clock Runtime}
\label{appendix:timing}

\paragraph{\textbf{Hardware and Setup}} We measured the wall-clock runtime for training and evaluation across the five longitudinal tasks (ILIADD, CHEEARS, ADNI, WOMAC, and KLG). To ensure a fair comparison, we ensured all methods had access to the same hardware: a single NVIDIA L40S GPU and an Intel Xeon Gold 6526Y CPU. To prevent timing mismatches caused by asynchronous CUDA executions, we synchronized the device before and after all timed regions and recorded the durations using \texttt{time.perf\_counter()}.

\begin{figure}[h]
     \centering
     \begin{subfigure}[b]{0.48\textwidth}
         \centering
         \includegraphics[width=\linewidth]{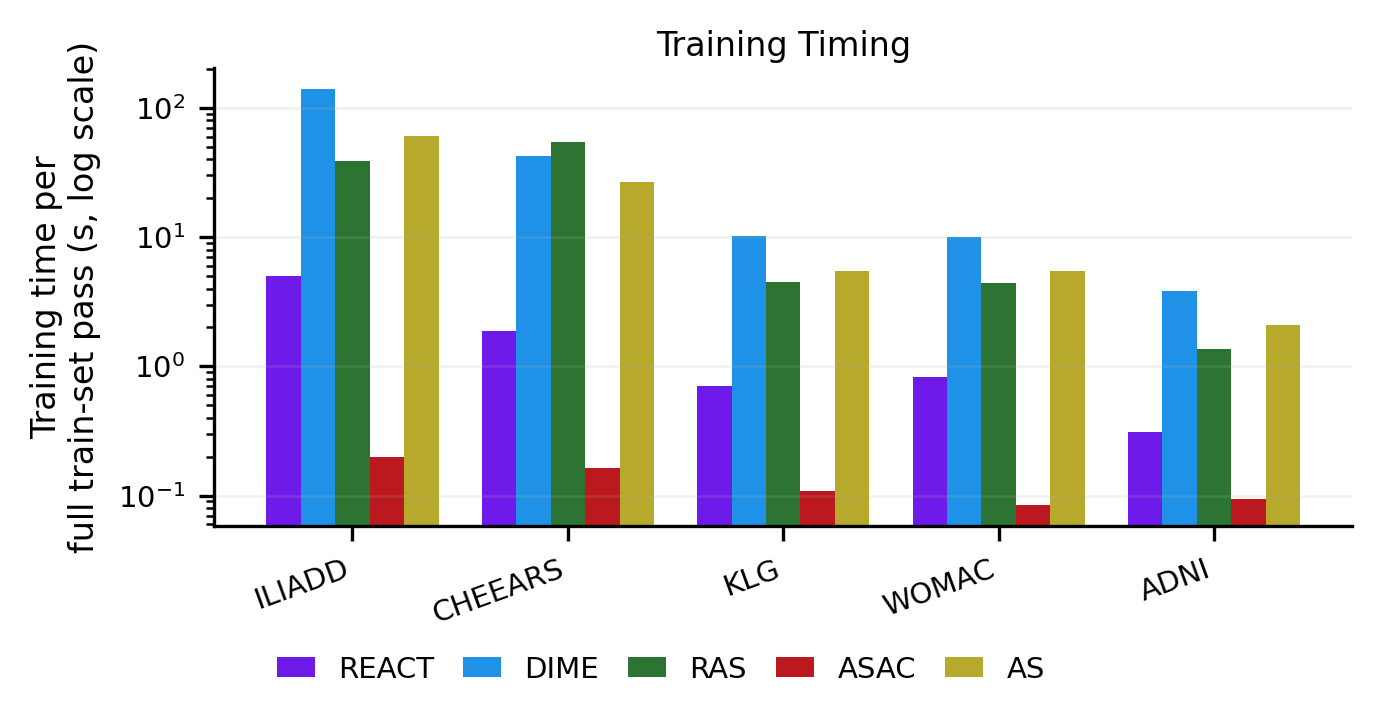}
         \caption{Training wall-clock runtime}
         \label{fig:training_time}
     \end{subfigure}
     \hfill
     \begin{subfigure}[b]{0.48\textwidth}
         \centering
         \includegraphics[width=\linewidth]{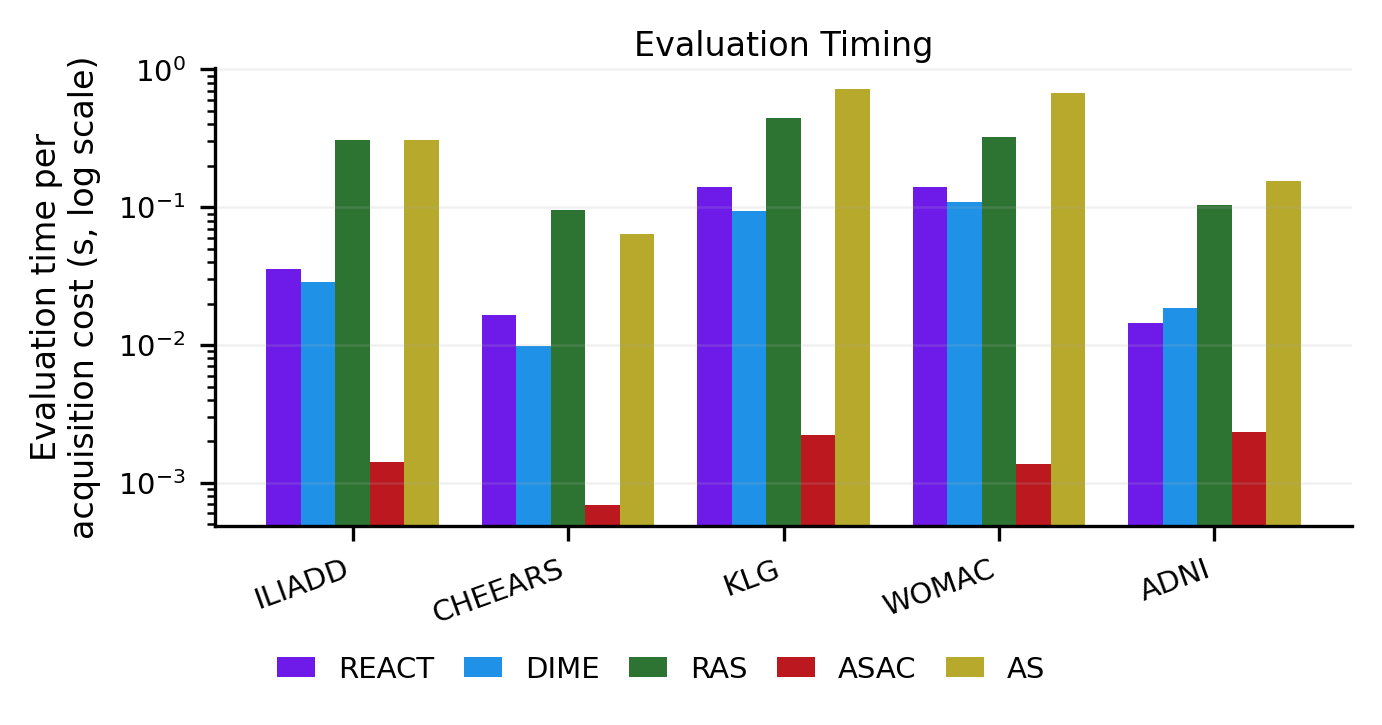}
         \caption{Evaluation wall-clock runtime}
         \label{fig:eval_time}
     \end{subfigure}
     \caption{Comparison of training and evaluation runtimes.}
     \label{fig:runtime_comparison}
\end{figure}

\paragraph{\textbf{Training Runtime}}
We normalized the training runtime as seconds per full train-set pass and used the same batch size during both training and evaluation. We define one iteration as one full train-set pass. The specific iterations for each method are 2 classifier pretraining iterations and 2 policy (joint) training iterations. In Fig.~\ref{fig:runtime_comparison}(a), we can see \OurMethod achieves training time that are faster than recent deep learning approaches (DIME, RAS, and AS).

\paragraph{\textbf{Evaluation Runtime}}
For inference, all methods were timed on the full test sets using the same batch size of 128. We report the end-to-end wall-clock runtime and the per-acquisition cost. In Fig.~\ref{fig:runtime_comparison}(b), we can see \OurMethod achieves comparable inference time with the baselines.

\section{Dataset Details}
\label{appendix:dataset}
\subsection{CHEEARS}
More details about features we used for the CHEEARS data can be found in \autoref{tab:cheears_context} and \autoref{tab:cheears_temporal}, see \cite{Ringwald2025CommonAU} for information on the features and survey design.

\begin{table*}[t]
\centering
\small
\begin{tabular}{|p{2.4cm}|p{1.4cm}|p{6cm}|}
\hline
\textbf{Feature Name} & \textbf{Acquisition Cost} & \textbf{Description} \\
\hline
sex & 1.0 & Participant sex. \\
\hline
age & 1.0 & Participant age. \\
\hline
race\_\_\_4 & 1.0 & Participant race (Black/African American indicator). \\
\hline
hispanic & 1.0 & Hispanic/Latino ethnicity. \\
\hline
education & 1.0 & Highest level of education attained. \\
\hline
current\_employed & 1.0 & Current employment status. \\
\hline
family\_income & 1.0 & Family income level. \\
\hline
religious\_affiliation & 1.0 & Religious affiliation. \\
\hline
cigarette\_use & 1.0 & Cigarette/tobacco use history. \\
\hline
alcohol\_use & 1.0 & Alcohol use history. \\
\hline
drug\_use & 1.0 & Drug use history. \\
\hline
mentalhealth & 1.0 & Mental health status or diagnosis. \\
\hline
audit\_total & 1.0 & Total score on the Alcohol Use Disorders Identification Test (AUDIT). \\
\hline
dmq\_soc & 1.0 & Drinking Motives Questionnaire (DMQ) --- Social motives subscale. \\
\hline
dmq\_cop & 1.0 & DMQ --- Coping motives subscale. \\
\hline
dmq\_enh & 1.0 & DMQ --- Enhancement motives subscale. \\
\hline
dmq\_con & 1.0 & DMQ --- Conformity motives subscale. \\
\hline
yaacq\_total & 1.0 & Young Adult Alcohol Consequences Questionnaire (YAACQ) --- total score. \\
\hline
yaacq\_social & 1.0 & YAACQ --- Social/interpersonal consequences subscale. \\
\hline
yaacq\_control & 1.0 & YAACQ --- Loss of control subscale. \\
\hline
yaacq\_selfperc & 1.0 & YAACQ --- Self-perception consequences subscale. \\
\hline
yaacq\_selfcare & 1.0 & YAACQ --- Self-care consequences subscale. \\
\hline
yaacq\_risk & 1.0 & YAACQ --- Risk/safety consequences subscale. \\
\hline
yaacq\_academic & 1.0 & YAACQ --- Academic/occupational consequences subscale. \\
\hline
yaacq\_depend & 1.0 & YAACQ --- Dependence symptoms subscale. \\
\hline
yaacq\_blackout & 1.0 & YAACQ --- Blackout subscale. \\
\hline
neo\_n & 1.0 & NEO Personality Inventory --- Neuroticism subscale. \\
\hline
neo\_e & 1.0 & NEO Personality Inventory --- Extraversion subscale. \\
\hline
neo\_a & 1.0 & NEO Personality Inventory --- Agreeableness subscale. \\
\hline
neo\_o & 1.0 & NEO Personality Inventory --- Openness subscale. \\
\hline
neo\_c & 1.0 & NEO Personality Inventory --- Conscientiousness subscale. \\
\hline
iip\_dom & 1.0 & Inventory of Interpersonal Problems (IIP) --- Dominance subscale. \\
\hline
iip\_lov & 1.0 & IIP --- Love/Affiliation subscale. \\
\hline
iip\_elev & 1.0 & IIP --- Elevation (overall interpersonal distress). \\
\hline
DoW & 1.0 & Day of the week for the first day of the sequence. \\
\hline
\end{tabular}

\caption{CHEEARS contextual descriptors.}
\label{tab:cheears_context}
More details about features we used for the ILIADD dataset can be found in \autoref{tab:iliadd_context} and \autoref{tab:iliadd_temporal}, see \cite{Ringwald2025CommonAU} for information on the features and survey design.

\end{table*}

\begin{table*}[t]
\centering
\small
\begin{tabular}{|p{2.4cm}|p{1.4cm}|p{6cm}|}
\hline
\textbf{Feature Name} & \textbf{Acquisition Cost} & \textbf{Description} \\
\hline
happy & 1.0 & Self-reported happiness in the past 15 minutes (continuous scale). \\
\hline
nervous & 1.0 & Self-reported nervousness in the past 15 minutes (continuous scale). \\
\hline
angry & 1.0 & Self-reported anger in the past 15 minutes (continuous scale). \\
\hline
sad & 1.0 & Self-reported sadness in the past 15 minutes (continuous scale). \\
\hline
excited & 1.0 & Self-reported excitement in the past 15 minutes (continuous scale). \\
\hline
alert & 1.0 & Self-reported alertness in the past 15 minutes (continuous scale). \\
\hline
ashamed & 1.0 & Self-reported shame in the past 15 minutes (continuous scale). \\
\hline
relaxed & 1.0 & Self-reported relaxation in the past 15 minutes (continuous scale). \\
\hline
bored & 1.0 & Self-reported boredom in the past 15 minutes (continuous scale). \\
\hline
content & 1.0 & Self-reported contentment in the past 15 minutes (continuous scale). \\
\hline
stress & 1.0 & Self-reported stress in the past 15 minutes (continuous scale). \\
\hline
drink\_plans & 1.0 & Whether participant has specific plans to drink tonight. \\
\hline
substance & 1.0 & Whether participant used any substances besides alcohol to get high or feel good. \\
\hline
dom & 1.0 & Self-rated dominance of social behavior during interactions today (continuous scale). \\
\hline
warm & 1.0 & Self-rated warmth of social behavior during interactions today (continuous scale). \\
\hline
drink\_likely & 1.0 & Likelihood of drinking tonight. \\
\hline
drink\_quantity & 1.0 & Anticipated number of drinks to consume tonight. \\
\hline
drink\_urge & 1.0 & Strength of urge to drink in the past 15 minutes. \\
\hline
nondrink\_likely & 1.0 & Likelihood of drinking tonight (non-drinking condition branch). \\
\hline
nondrink\_quantity & 1.0 & Anticipated drink quantity (non-drinking condition branch). \\
\hline
nondrink\_urge & 1.0 & Urge to drink in the past 15 minutes (non-drinking condition branch). \\
\hline
nondrink\_plan\_other & 1.0 & Free-text specification of other evening plans (non-drinking branch). \\
\hline
daily\_activities & 1.0 & Activities completed today (9 binary features). \\
\hline
daily\_experiences & 1.0 & Work-related experiences today (6 binary features). \\
\hline
drink\_expectancies & 1.0 & Expected outcomes if drinking tonight (29 binary features). \\
\hline
drink\_motives & 1.0 & Reasons for drinking tonight (13 binary features). \\
\hline
general\_experiences & 1.0 & General experiences that occurred today (4 binary features). \\
\hline
nondrink\_expectancies & 1.0 & Expected outcomes for tonight if not drinking (26 binary features). \\
\hline
nondrink\_motives & 1.0 & Drinking motives, non-drinking condition branch (13 binary features). \\
\hline
nondrink\_plans & 1.0 & Plans for the evening if not drinking (13 binary features). \\
\hline
social\_experiences & 1.0 & Social experiences that occurred today (7 binary features). \\
\hline
\end{tabular}
\caption{CHEEARS temporal features.}
\label{tab:cheears_temporal}
\end{table*}
\subsection{ILIADD}
More details about features we used for the ILIADD data can be found in \autoref{tab:iliadd_context} and \autoref{tab:iliadd_temporal}.

\begin{table*}[t]
\centering
\small
\begin{tabular}{|p{2.4cm}|p{1.4cm}|p{6cm}|}
\hline
\textbf{Feature Name} & \textbf{Acquisition Cost} & \textbf{Description} \\
\hline
age & 1.0 & Participant age. \\
\hline
sex & 1.0 & Participant sex. \\
\hline
handedness & 1.0 & Participant handedness (e.g., left, right, ambidextrous). \\
\hline
multiracial & 1.0 & Multiracial identity indicator. \\
\hline
hispanic & 1.0 & Hispanic/Latino ethnicity. \\
\hline
language & 1.0 & Primary language spoken. \\
\hline
marital & 1.0 & Marital status. \\
\hline
relationship & 1.0 & Current romantic relationship status. \\
\hline
grade & 1.0 & Current academic grade level. \\
\hline
degree & 1.0 & Degree program or highest degree attained. \\
\hline
income & 1.0 & Family or personal income level. \\
\hline
cigarette & 1.0 & Cigarette/tobacco use. \\
\hline
substance & 1.0 & Use of substances other than alcohol to get high or feel good. \\
\hline
treatment & 1.0 & History of mental health or substance use treatment. \\
\hline
recentTreatment & 1.0 & Whether participant received treatment recently. \\
\hline
whoTreatment & 1.0 & Type or provider of most recent treatment. \\
\hline
HiTOP\_Dishon & 1.0 & HiTOP --- Disinhibition/Dishonesty spectrum score. \\
\hline
HiTOP\_DisDys & 1.0 & HiTOP --- Disinhibited/Dysregulated spectrum score. \\
\hline
HiTOP\_Emot & 1.0 & HiTOP --- Emotional Dysfunction spectrum score. \\
\hline
HiTOP\_Mistrust & 1.0 & HiTOP --- Mistrust/Antagonism spectrum score. \\
\hline
HiTOP\_PhobInd & 1.0 & HiTOP --- Phobic Internalizing spectrum score. \\
\hline
BFI\_E & 1.0 & Big Five Inventory --- Extraversion subscale. \\
\hline
BFI\_A & 1.0 & Big Five Inventory --- Agreeableness subscale. \\
\hline
BFI\_C & 1.0 & Big Five Inventory --- Conscientiousness subscale. \\
\hline
BFI\_N & 1.0 & Big Five Inventory --- Neuroticism subscale. \\
\hline
BFI\_O & 1.0 & Big Five Inventory --- Openness to Experience subscale. \\
\hline
\end{tabular}
\caption{ILIADD contextual descriptors.}
\label{tab:iliadd_context}
\end{table*}

\begin{table*}[t]
\centering
\small
\begin{tabular}{|p{2.4cm}|p{1.4cm}|p{6cm}|}
\hline
\textbf{Feature Name} & \textbf{Acquisition Cost} & \textbf{Description} \\
\hline
interaction & 1.0 & Event-contingent report of a social interaction (triggered by participant). \\
\hline
positiveaffEMA & 1.0 & Self-reported positive affect in the past 15 minutes (0 = Neutral, 10 = Very Positive). \\
\hline
energyEMA & 1.0 & Self-reported energy/alertness in the past 15 minutes (0 = Not at all, 10 = Very Energetic/Awake). \\
\hline
stressEMA & 1.0 & Self-reported stress in the past 15 minutes (0 = Not at all, 10 = Extremely). \\
\hline
impulse1EMA & 1.0 & In the past hour, acted on impulse (0 = Not at all, 10 = Very Much). \\
\hline
impulse2EMA & 1.0 & In the past hour, did things without worrying about consequences (0 = Not at all, 10 = Very Much). \\
\hline
impulse3EMA & 1.0 & In the past hour, decided to put off something that needed to be done (0 = Not at all, 10 = Very Much). \\
\hline
impulse4EMA & 1.0 & In the past hour, avoided doing something despite knowing the consequences (0 = Not at all, 10 = Very Much). \\
\hline
\end{tabular}
\caption{ILIADD temporal (EMA) features.}
\label{tab:iliadd_temporal}
\end{table*}
\subsection{ADNI}
More details about the ADNI context descriptors and temporal features could be found in \autoref{tab:adni_context} and \autoref{tab:adni_temporal}, respectively. Data access and the corresponding Data Use Agreement can be found at \url{https://adni.loni.usc.edu} and \url{https://adni.loni.usc.edu/terms-of-use/}, respectively.

\begin{table*}[t]
\centering
\small
\begin{tabular}{|p{2.4cm}|p{1.4cm}|p{6cm}|}
\hline
\textbf{Feature Name} & \textbf{Acquisition Cost} & \textbf{Description} \\
\hline
AGE & 0.3 & Participant age. \\
\hline
PTGENDER & 0.3 & Participant gender. \\
\hline
PTEDUCAT & 0.3 & Participant education. \\
\hline
PTETHCAT & 0.3 & Participant ethnic category. \\
\hline
PTRACCAT & 0.3 & Participant racial category. \\
\hline
PTMARRY & 0.3 & Participant marital status. \\
\hline
FAQ & 0.3 & Functional Activities Questionnaire total score. \\
\hline
\end{tabular}
\caption{ADNI contextual descriptors.}
\label{tab:adni_context}
\end{table*}

\begin{table*}[t]
\centering
\small
\begin{tabular}{|p{2.4cm}|p{1.4cm}|p{6cm}|}
\hline
\textbf{Feature Name} & \textbf{Acquisition Cost} & \textbf{Description} \\
\hline
FDG & 1.0 & FDG PET biomarker. \\
\hline
AV45 & 1.0 & AV45 PET biomarker. \\
\hline
Hippocampus & 0.5 & Hippocampal MRI biomarker. \\
\hline
Entorhinal & 0.5 & Entorhinal MRI biomarker. \\
\hline
\end{tabular}
\caption{ADNI temporal features.}
\label{tab:adni_temporal}
\end{table*}

\subsection{OAI}
The OAI dataset involves two prediction tasks, WOMAC and KLG, which share the same set of input features. More details about the OAI context descriptors and temporal features can be found in \autoref{tab:oai_context} and \autoref{tab:oai_temporal}, respectively. Access to the OAI data may be requested via \url{https://nda.nih.gov/oai/}.

\begin{table*}[t]
\centering
\small
\begin{tabular}{|p{2.4cm}|p{1.4cm}|p{6cm}|}
\hline
\textbf{Feature Name} & \textbf{Acquisition Cost} & \textbf{Description} \\
\hline
HISP & 0.3 & Hispanic/Latino ethnicity indicator. \\
\hline
RACE & 0.3 & Self-reported race category. \\
\hline
SEX & 0.3 & Participant sex. \\
\hline
FAMHXKR & 0.3 & Mother, father, sister, or brother had knee repl surgery where all/part of knee replaced. \\
\hline
EDCV & 0.3 & Highest grade or year of school completed. \\
\hline
AGE & 0.3 & Age. \\
\hline
SMOKE & 0.3 & Smoking history/status. \\
\hline
INCOME2 & 0.3 & Household income category. \\
\hline
MARITST & 0.3 & Marital status. \\
\hline
MEDINS & 0.3 & Medical / health insurance status. \\
\hline
\end{tabular}
\caption{OAI contextual descriptors.}
\label{tab:oai_context}
\end{table*}

\begin{table*}[t]
\centering
\small
\begin{tabular}{|p{2.4cm}|p{1.4cm}|p{6cm}|}
\hline
\textbf{Feature Name} & \textbf{Acquisition Cost} & \textbf{Description} \\
\hline
DRNKAMT & 0.3 & Current alcohol consumption amount. \\
\hline
DRKMORE & 0.3 & Alcohol-use indicator capturing heavier or more frequent drinking pattern. \\
\hline
BPSYS & 0.5 & Systolic blood pressure. \\
\hline
BPDIAS & 0.5 & Diastolic blood pressure. \\
\hline
BMI & 0.5 & Body mass index. \\
\hline
CEMPLOY & 0.3 & Current employment. \\
\hline
CUREMP & 0.3 & Currently work for pay. \\
\hline
JSW\_1 $\rightarrow$ JSW\_10  & 0.8 & Fixed-location radiographic knee joint space width (JSW) measurements. \\
\hline
\end{tabular}
\caption{OAI temporal features.}
\label{tab:oai_temporal}
\end{table*}

\section{Onboarding Context Selection}
\label{appendix:onboarding_context}
\begin{figure*}[p]
    \centering
    \begin{subfigure}[b]{0.7\textwidth}
        \centering
        \includegraphics[width=\textwidth, trim={0 6cm 0 4cm}, clip]{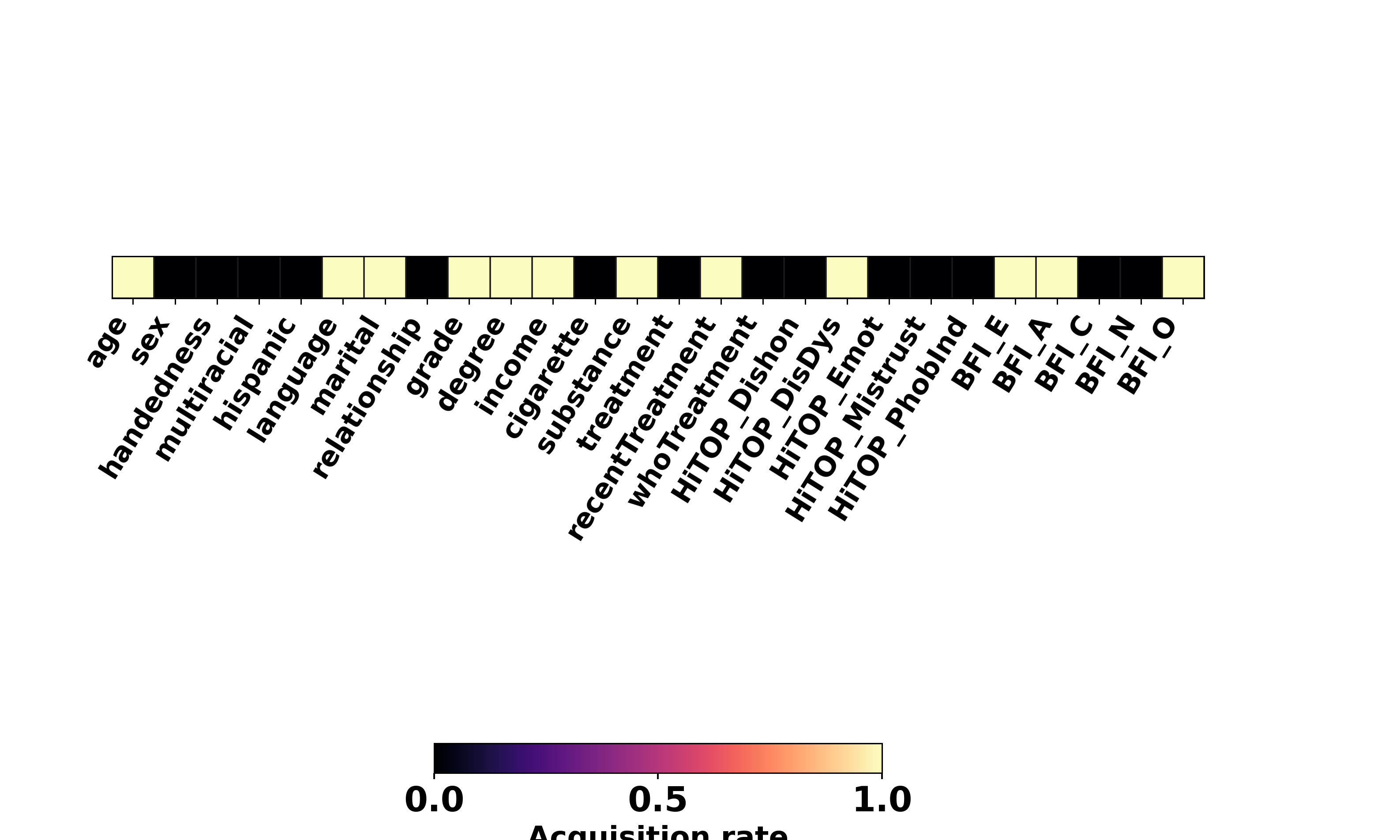}
        \caption{ILIADD}
    \end{subfigure}
    
    \begin{subfigure}[b]{0.8\textwidth}
        \centering
        \includegraphics[width=\textwidth, trim={0 5.5cm 0 4cm}, clip]{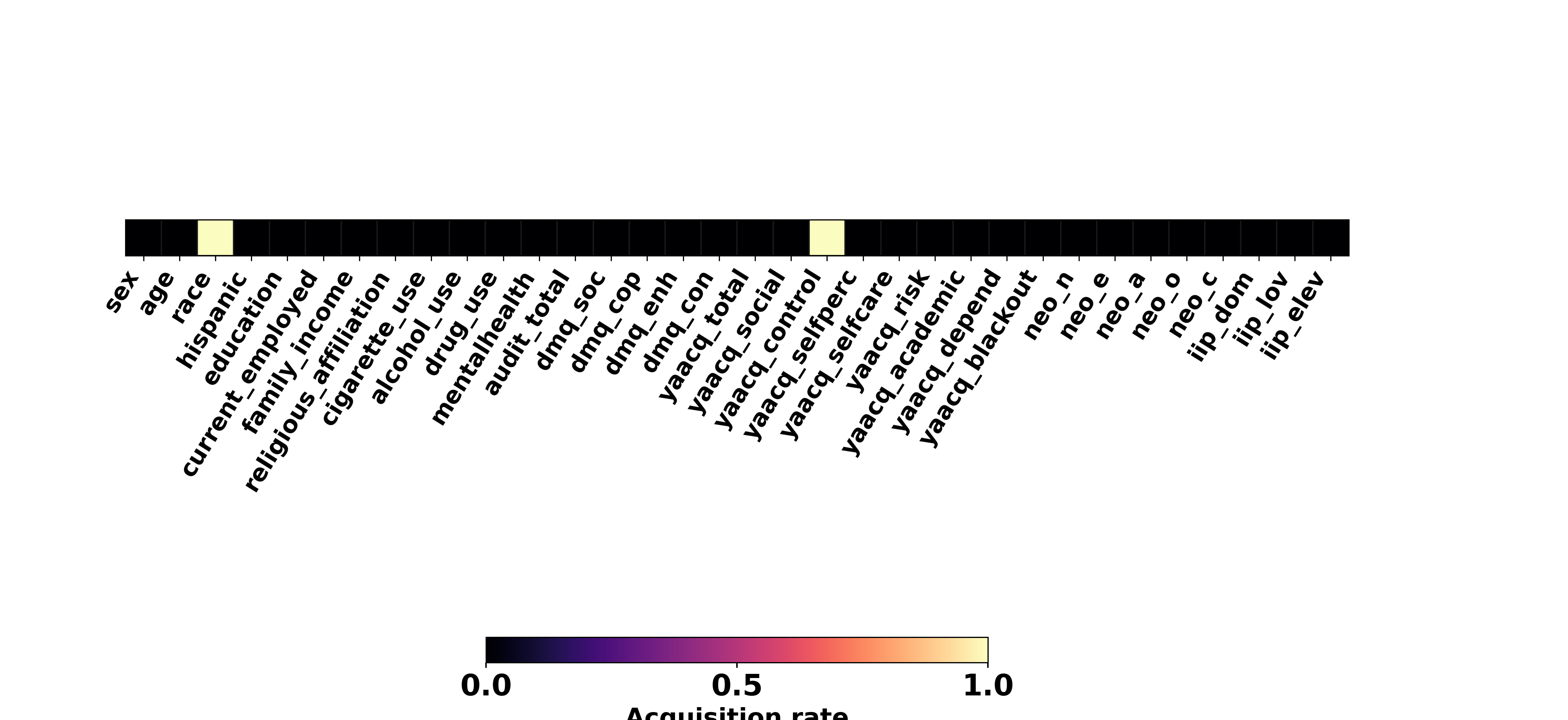}
        \caption{CHEEARS}
    \end{subfigure}

    \begin{subfigure}[b]{0.5\textwidth}
        \centering
        \includegraphics[width=\textwidth, trim={0 6.5cm 0 3cm}, clip]{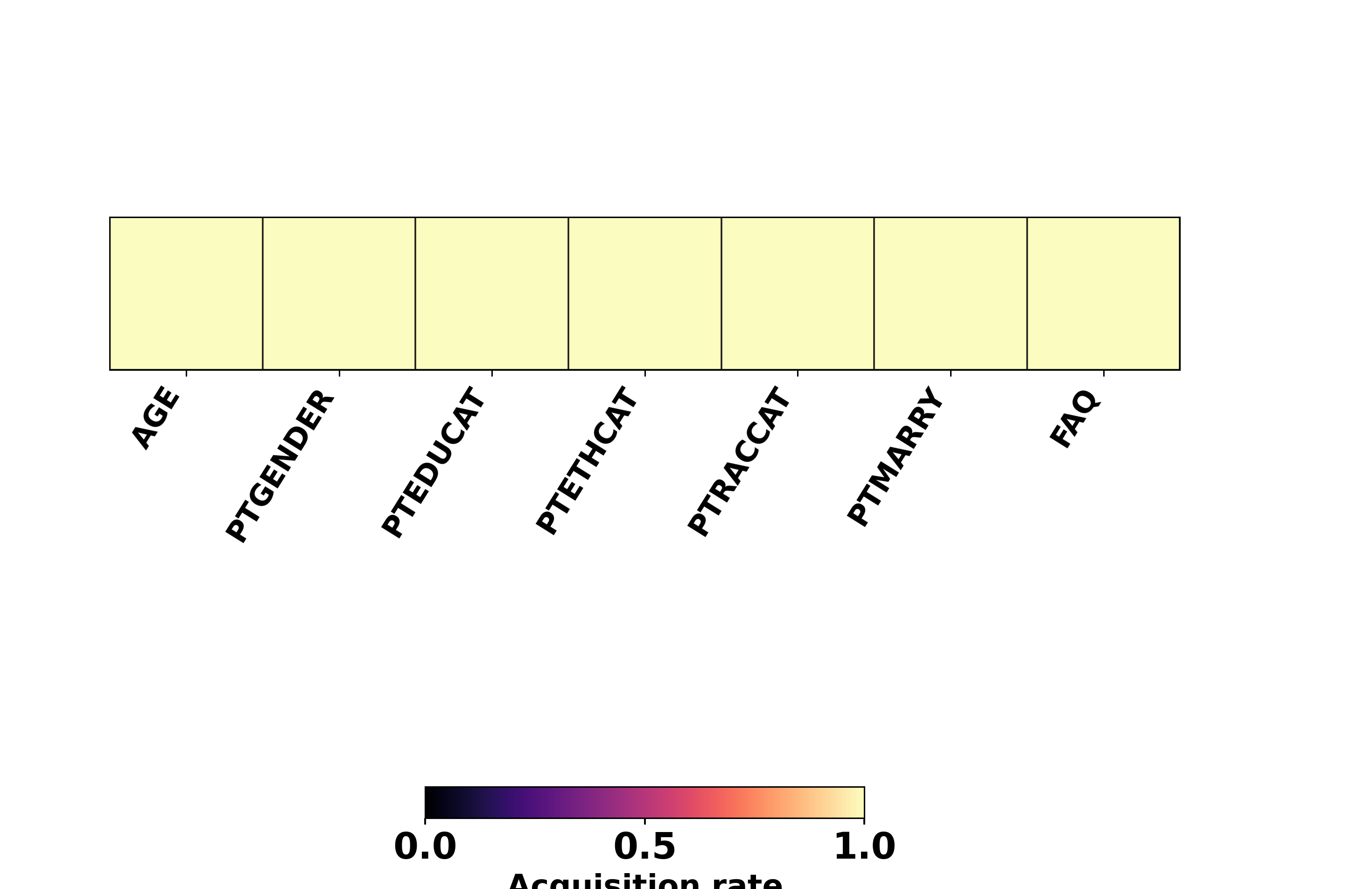}
        \caption{ADNI}
    \end{subfigure}

    \begin{subfigure}[b]{0.6\textwidth}
        \centering
        \includegraphics[width=\textwidth, trim={0 7cm 0 4cm}, clip]{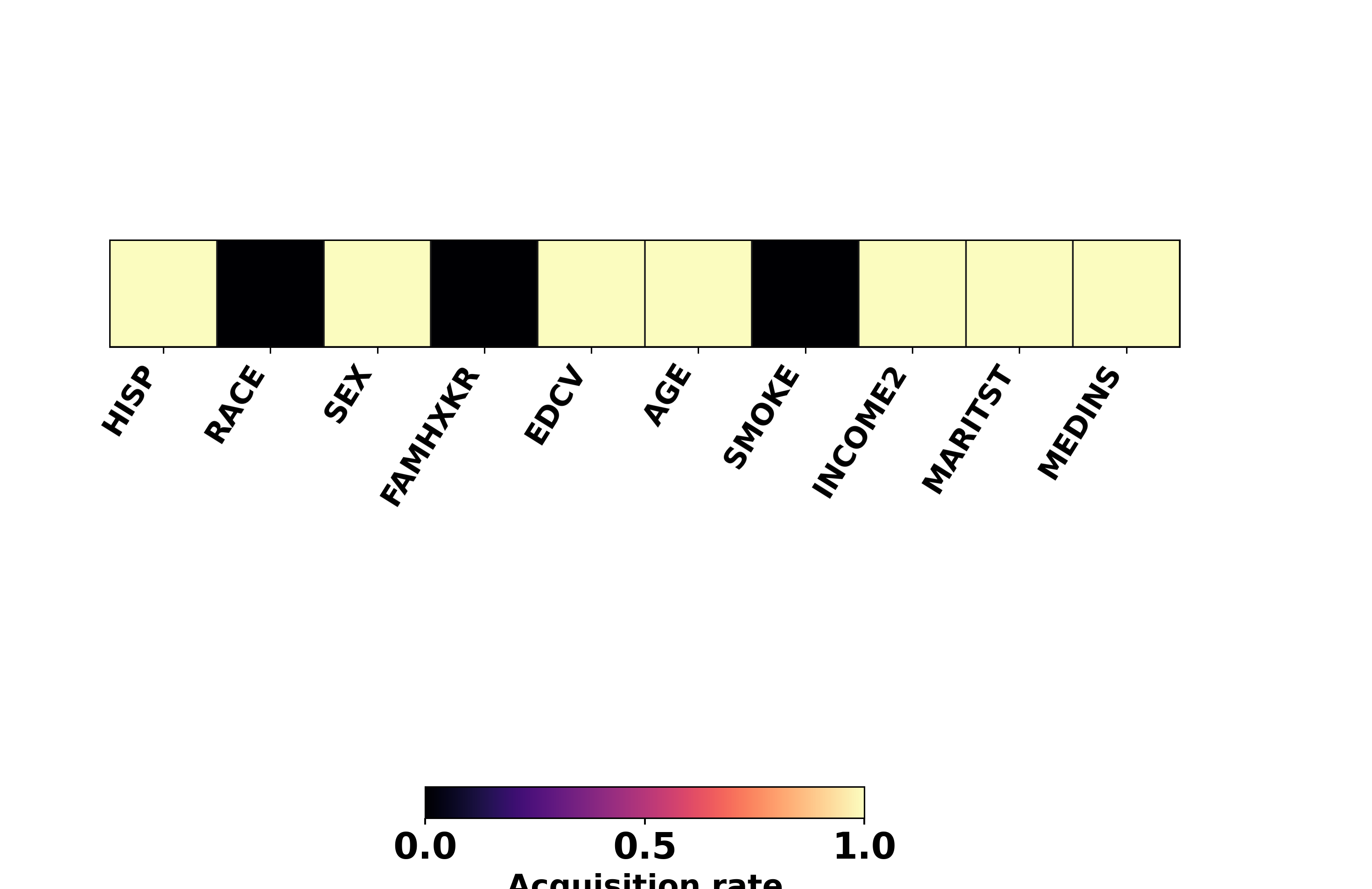}
        \caption{WOMAC}
    \end{subfigure}

    \begin{subfigure}[b]{0.6\textwidth}
        \centering
        \includegraphics[width=\textwidth, trim={0 7cm 0 4cm}, clip]{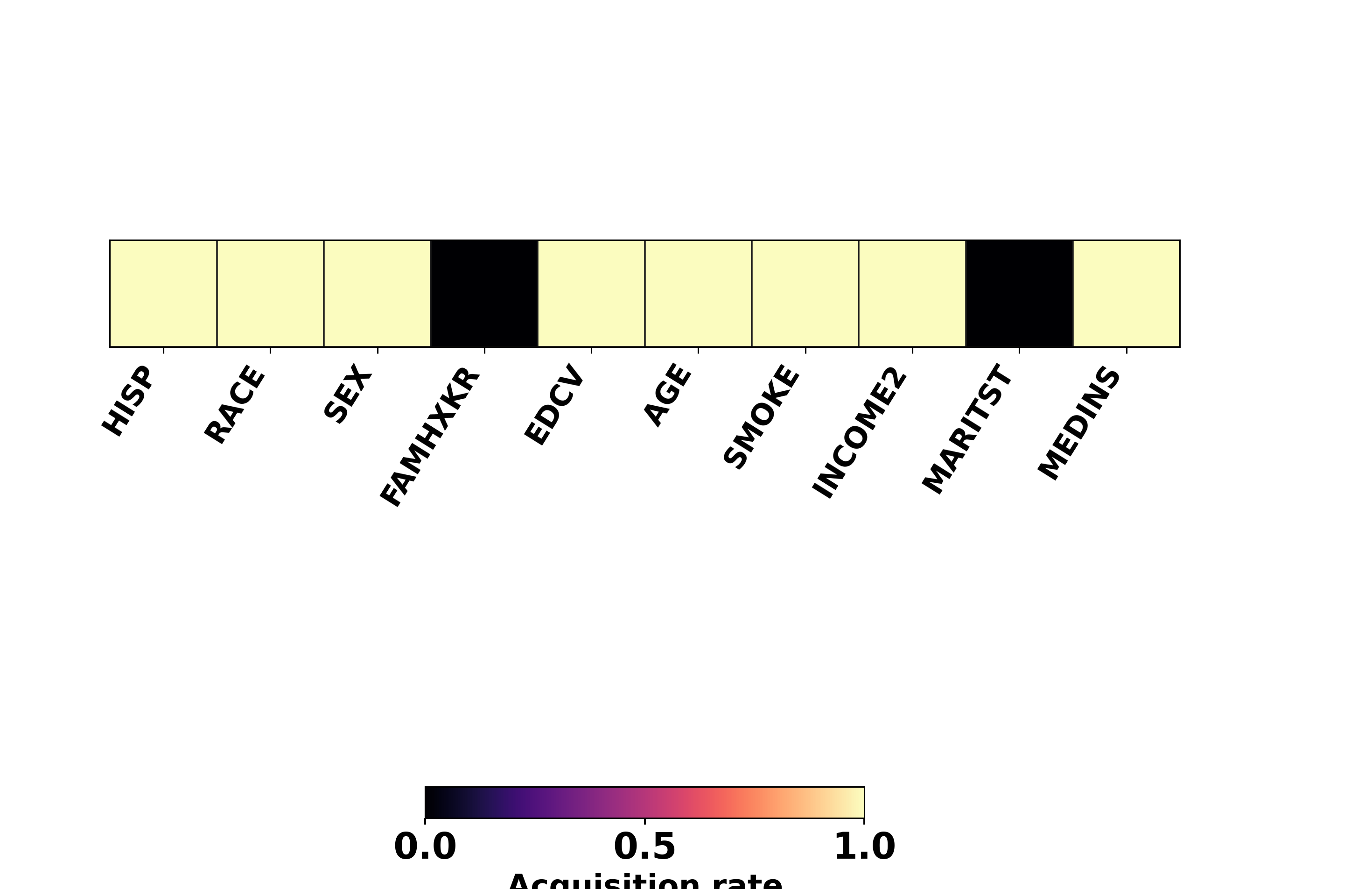}
        \caption{KLG}
    \end{subfigure}
    
    \caption{Selected contexts for the policies shown in Fig.~\ref{fig:traj_all}. Bright cells indicate selected features, while dark cells indicate those that were not selected.}
    \label{fig:learned_contexts}
\end{figure*}

We show examples of the selected contexts in Fig.~\ref{fig:learned_contexts}. These contexts correspond to the policies illustrated in Fig.~\ref{fig:traj_all}.

\end{document}